\documentclass{article}
\usepackage[preprint]{neurips_2023}

\usepackage[utf8]{inputenc} 
\usepackage[T1]{fontenc}    
\usepackage{hyperref}       
\usepackage{url}            
\usepackage{booktabs}       
\usepackage{amsfonts}       
\usepackage{nicefrac}       
\usepackage{microtype}      
\usepackage{xcolor}         
\usepackage{ulem}  
\usepackage{float}
\usepackage{graphicx} 
\usepackage{epstopdf}
\usepackage{amsmath}
\usepackage{times}
\usepackage{latexsym}
\usepackage{multirow}
\usepackage{multicol}
\usepackage{subfigure}
\usepackage{diagbox}
\usepackage{makecell}
\usepackage{graphics}
\usepackage{amsfonts,amssymb} 
\usepackage{booktabs}
\usepackage{tabularx}
\usepackage{enumitem}
\usepackage{array}
\usepackage{listings} 
\usepackage{adjustbox}  
\usepackage{wrapfig}

\usepackage{color}
\usepackage{mdframed}
\usepackage{CJKutf8}
\newmdenv[linecolor=black, backgroundcolor=blue!5]{questionbox}
\newmdenv[linecolor=black, backgroundcolor=red!5]{answerbox}
\newmdenv[linecolor=black, backgroundcolor=white!5]{sample}
\title{\raisebox{-0.10cm}{\includegraphics[scale=0.08]{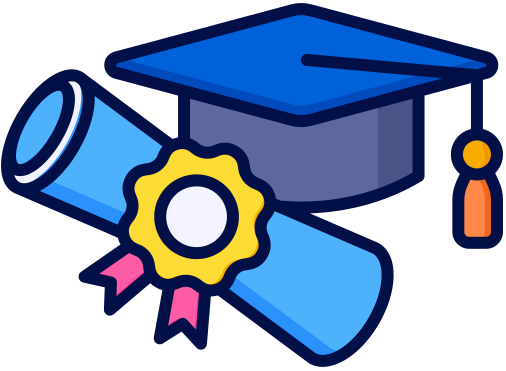}} \textcolor[rgb]{0.965,0.325,0.078}{Open}\textcolor[rgb]{0,0.631,0.945}{B}\textcolor[rgb]{1,0.733,0}{A}: An \textcolor[rgb]{0.965,0.325,0.078}{Open}-Sourced 15B \textcolor[rgb]{0,0.631,0.945}{B}ilingual \textcolor[rgb]{1,0.733,0}{A}symmetric Seq2Seq Model Pre-trained from Scratch}

\author{ Juntao Li\thanks{\; Project Leader. \url{ljt@suda.edu.cn}}, \ Zecheng Tang$^\dagger$, \ Yuyang Ding$^\dagger$, \ Pinzheng Wang\thanks{\; Equal Contribution. \url{{zctang,yyding23,pzwang1}@stu.suda.edu.cn}}  \\
\\
\textbf{Pei Guo}, \ \textbf{Wangjie You}, \ \textbf{Dan Qiao},\ \textbf{Wenliang Chen},\ \textbf{Guohong Fu}, \ \textbf{Qiaoming Zhu}, \\
\\
\textbf{Guodong Zhou}$^\ddagger$, \ \textbf{Min Zhang}\thanks{\; Corresponding Author: \url{{gdzhou,minzhang}@suda.edu.cn}} \\
\AND
{\fontsize{11pt}{12pt}\selectfont \textbf{Soochow University}}
}

\begin{document}

\textcolor{blue}{\textit{Update: We have performed full-parameter fine-tuning with specialized datasets, enabling OpenBA to become the expert model (OpenBA-X) for downstream tasks~(Bilingual Multi-turn Dialogue, Code Generation, Instruction Generation, and Tool Retrieval).}}
\maketitle

\begin{abstract}

Large language models (LLMs) with billions of parameters have demonstrated outstanding performance on various natural language processing tasks.
This report presents OpenBA, an open-sourced 15B bilingual asymmetric seq2seq model, to contribute an LLM variant to the Chinese-oriented open-source model community.
We enhance OpenBA with effective and efficient techniques as well as adopt a three-stage training strategy to train the model from scratch.
Our solution can also achieve very competitive performance with only 380B tokens, which is better than LLaMA-70B on the BELEBELE benchmark, BLOOM-176B on the MMLU benchmark, GLM-130B on the C-Eval (hard) benchmark.
This report provides the main details to pre-train an analogous model, including pre-training data processing, Bilingual Flan data collection, the empirical observations that inspire our model architecture design, training objectives of different stages, and other enhancement techniques.
Additionally, we also provide the fine-tuning details of OpenBA on four downstream tasks.
We have refactored our code to follow the design principles of the Huggingface Transformers Library, making it more convenient for developers to use, and released checkpoints of different training stages at \url{https://huggingface.co/openBA}. More details of our project are available at \url{https://github.com/OpenNLG/openBA.git}.
\end{abstract}

\clearpage
\tableofcontents 
\clearpage
\section{Introduction}
\label{sec:introduction}

The scaling law~\citep{kaplan2020scaling,clark2022unified,hoffmann2022training,touvron2023llama} of language models has brought unprecedented success. These large language models pre-trained on massive textual data demonstrate enormous superiority over previous paradigms for various fields and even have obtained newly emergent capabilities.
Though very powerful and developing rapidly, these models at scale are still far from perfect or satisfactory for most of the real-world usages. 
To advance the development of LLMs, the open-source community has made great efforts to provide strong and publicly accessible LLMs, covering different data sources, architectures, language modeling objectives, training pipelines, model scales, and language of expertise, e.g., BLOOM~\citep{scao2022bloom}, LLaMA~\citep{touvron2023llama,touvron2023llama2}, FlanT5~\citep{chung2022scaling}, AlexaTM~\citep{soltan2022alexatm}.

As for Chinese, the open-source community has also released many large language models either by pre-training from scratch, e.g., GLM~\citep{zeng2022glm}, Baichuan~\citep{Baichuan2023} or conducting further fine-tuning on existing open-sourced multilingual models, e.g., Huatuo~\citep{wang2023huatuo}, Luotuo~\citep{luotuo}, Phoenix~\citep{chen2023phoenix}, Chinese-LLaMA~\citep{cui2023efficient}, MOSS~\citep{sun2023moss}. These publicly available LLMs provide researchers and developers with strong general language models (i.e., the framework used by GLM~\citep{du2022glm}) and different decoder-only variants, but leaving the Encoder-Decoder framework (e.g., Flan-T5~\citep{chung2022scaling}) under-explored, which has been proven universally effective for different prompt settings (zero-shot, few-shot, and chain-of-thought)~\citep{longpre2023flan} and various tasks (e.g., language understanding, commonsense reasoning, question answering, information retrieval, and multi-turn chit-chat conversation)~\citep{tay2022ul2,zheng2023judging}.

To fill this blank, we contribute an open-sourced 15B bilingual asymmetric seq2seq model (OpenBA) pre-trained from scratch, providing not only the model checkpoints but also the data collection and processing details to construct pre-training data and bilingual Flan collection from openly accessible data resources (e.g., Common Crawl, the Pile corpus, C-Book), the motivations and empirical observations for the model architecture design, and the key details of other enhancement strategies. 
Concretely, our collected pre-training data consists of balanced English and Chinese tokens to make the Chinese language modeling benefit from high-quality English data. 
Since it is difficult to construct a Flan-like Chinese collection that covers extensive tasks and settings from open resources, we incorporate more English data sampled from the Flan collection in our Bilingual-Flan corpus. 
Unlike the vanilla Flan-T5~\citep{chung2022scaling} of a balanced encoder-decoder structure and the asymmetric deep-encoder shallow-decoder in AlexaTM~\citep{soltan2022alexatm}, we utilize another asymmetric model structure, i.e., shallow-encoder deep-decoder to enhance the generation capability, which is motivated by our empirical study in Sec.~\ref{subsec:model_arch_selection}.
Our training process comprises three different stages, including the UL2 pre-training, length-adaptation, and Flan training. 
We also incorporate enhancement strategies in model architecture and training to improve model capability, training stability, and efficiency. 
Evaluations across different benchmarks (MMLU, CMMLU, C-Eval, SuperGLUE, BELEBELE, BBH) and tasks (e.g., understanding, reasoning, and generation) have demonstrated the effectiveness of our model in different settings (zero-shot, few-shot, held-in, and held-out).  
Though merely trained on 380B tokens, our model can surpass many representative models, e.g., LLaMA-70B on BELEBELE, BLOOM-176B on MMLU, ChatGLM-6B on CMMLU and C-Eval.
Throughout the whole training process, OpenBA-15B produces around \textit{6.5 $tCO_{2eq}$ in total},
which is much less than the LLaMA-7B model that costs \textit{14 $tCO_{2eq}$}.

Additionally, we further fine-tune the model on four downstream tasks, including bilingual multi-turn dialogue, code generation, instruction generation, and tool retrieval, enabling OpenBA to become the expert model (OpenBA-X) for the downstream tasks.
All the implementation details are open-accessible, not limited to data collection and processing, codes, model checkpoints, and evaluations.
As we are still working on a few directions to improve and apply the OpenBA model, we welcome any comments and suggestions and look forward to further cooperation with the open-source community.

\section{Related Work}
\label{sec:related_work}

\subsection{Large Language Models}
Benefiting from scaling law~\citep{kaplan2020scaling,clark2022unified,hoffmann2022training} and the growth of computational resources~\citep{schaller1997moore}, the recent years have witnessed the remarkable evolution in the field of LLMs, which pushes the boundaries of various NLP tasks~\citep{radford2018improving,brown2020language,zeng2021pangu,sun2021ernie,zhang2021commentary,zhang2021cpm,zhang2022opt,touvron2023llama}. 
The introduction of transformer model~\citep{vaswani2017attention} is a notable turning point in the NLP field, as models based on such an architecture like GPT-2~\citep{radford2019language} and T5~\citep{raffel2020exploring} have demonstrated outstanding performance across a wide range of NLP tasks by unifying the formulation of different tasks and scaling up the model size.
This trend has continued with the advent of GPT-3~\citep{brown2020language}, which brings about groundbreaking advancements in scaling with its 175-billion-parameter model. 
Consequently, the research community has gradually recognized the benefits of LLMs, leading to a series of subsequent models following in rapid succession, such as Gopher~\citep{rae2021scaling}, Megatron-Turing~\citep{smith2022using}, Chinchilla~\citep{hoffmann2022training}, BLOOM~\citep{scao2022bloom}, LLaMA~\citep{touvron2023llama,touvron2023llama2}, ChatGPT~\citep{ouyang2022training,bubeck2023sparks}, Falcon~\citep{penedo2023refinedweb}, etc.
These models have consistently advanced the frontiers in the NLP domain, promoting ongoing development and progress.
However, in this process, two serious issues have gradually emerged. 
The first issue is the open-sourcing of LLMs. 
Due to concerns such as data sources and privacy~\citep{pan2020privacy}, many LLMs are not available to the public or can only be accessed via limited or commercial APIs, e.g., ChatGPT~\citep{ouyang2022training} and PaLM~\citep{chowdhery2022palm}, while the open-source alternatives have relative lower performance compared to their closed-source counterparts~\citep{hendrycks2020measuring,li2023cmmlu}.
Another issue is that, following the success of decoder-only models like GPT-3 and ChatGPT, the current focus of research mainly revolves around decoder-only architecture, while the investigation on large-scale encoder-decoder models, such as T5~\citep{raffel2020exploring} and AlexaTM~\citep{soltan2022alexatm}, presents a relatively ``under-explored area'' in this field.
Additionally, there is no clear consensus on whether encoder-decoder or decoder-only models hold an advantage over the others in terms of architectural superiority~\citep{tay2022ul2,fu2023decoder}.
In an effort to contribute to the open-source community and supplement the existing encoder-decoder models, we developed OpenBA, featuring an asymmetric encoder-decoder architecture.
We collect and filter the pre-training data from open-accessible corpora.
Additionally, we manually construct the Chinese Flan-like data from various publicly available annotated datasets and combine them with the English Flan corpus to obtain the Bilingual Flan collection.
We employ a stage-wise training strategy to optimize the model performance by utilizing these datasets.
Our model achieves outstanding performance on multiple widely-used benchmarks, such as SuperGLUE~\citep{wang2019superglue} and C-Eval~\citep{huang2023c}.

\subsection{Instruction Tuning}
Instruction tuning, which involves the method of fine-tuning LLMs on an instruction dataset in a supervised manner, has played a crucial role in the significant advancements of LLMs in recent years~\citep{zhang2023instruction}.
Starting from the T5 model~\citep{raffel2020exploring}, which pioneers the concept of consolidating diverse NLP tasks as generative tasks.
By employing task-specific prompts to guide the model, this method streamlines the process of applying LLMs to an extensive array of applications, laying the foundation for subsequent instruction tuning models such as FLAN~\citep{wei2021finetuned,chung2022scaling} and T0~\citep{sanh2021multitask}, which further improve performance across diverse tasks by incorporating more task-specific instructions during the pre-training phase.
An approach related to instruction tuning is chain-of-thought~(CoT) prompting~\citep{nye2021show,wei2022chain}, which enhances instructions with descriptions of intermediate reasoning steps, thereby boosting LLM performance~\citep{wang2022self,zelikman2022star,wu2023improving,xu2023wizardlm}.
At present, the open-source community offers a multitude of instruction datasets, such as Alpaca~\citep{alpaca} and Dolly~\citep{conover2023free}. 
These instructions aim to enhance specific professional abilities of LLMs, such as code generation ability~\citep{codealpaca}, or the general capabilities like commonsense reasoning skills~\citep{zhang2023automatic}. 
However, the wide variety and inconsistent quality of these datasets pose challenges, with each dataset typically comprising a relatively small amount of data and focusing on a single language.
In this work, we construct the BiFlan dataset, the first Bilingual Flan dataset built upon the cleansed Flan data~\citep{longpre2023flan}, containing various instruction types and tasks in English and Chinese language. 
Experimental results indicate that training on the BiFlan dataset can significantly enhance model performance on various strong benchmarks, such as  MMLU~\citep{hendrycks2020measuring} and CMMLU~\citep{li2023cmmlu}.

\section{Methodology}
\label{sec:approach}
This section presents the details of our OpenBA model, including our considerations and implementations in pre-training data collection and processing, Bilingual Flan data construction, model architecture, training objectives and pipelines, as well as the model implementation and techniques.

\subsection{Dataset Collection}
\label{sec:dataset_collection}
This part elaborates on the data collection and filtering process for each training stage: pre-training data for stage \uppercase\expandafter{\romannumeral1} and \uppercase\expandafter{\romannumeral2}~(Sec.~\ref{subsec:pretrain_data_filtering}), and Bilingual Flan~(BiFlan) data for stage \uppercase\expandafter{\romannumeral3}~(Sec.~\ref{subsec:bilingual_flan_data_collection}).

\subsubsection{Pre-training Data Collection and Filtration}
\label{subsec:pretrain_data_filtering}

\paragraph{Data Sources}
Considering that our primary target is to construct an open-source model, we collect our pre-training data from publicly accessible resources consisting of English, Chinese, and code data. 
Concretely, the English and code data are sampled from the Pile corpus~\citep{gao2020pile}\footnote{\url{https://pile.eleuther.ai/}}, which is a collection of 22 diverse high-quality subsets.
The Chinese data is mainly collected from the Internet (i.e., a cleaned subset from Common Crawl\footnote{\url{https://commoncrawl.org/}}), Chinese books (i.e., C-Book\footnote{\url{https://github.com/FudanNLPLAB/CBook-150K}}), News (i.e., news2016zh\footnote{\label{foot:clue}\url{https://github.com/CLUEbenchmark/CLUE}}), Encyclopedias (i.e., baidu\_baike\footnote{\url{https://github.com/BIT-ENGD/baidu_baike}} and wiki2019zh\_corpus\textsuperscript{\ref{foot:clue}}), Comments (i.e., comments2019zh\_corpus\textsuperscript{\ref{foot:clue}}) and Laws (i.e., CAIL2018\footnote{\url{https://github.com/thunlp/CAIL}}).

\paragraph{Filtration}
Before mixing these different data components with a certain proportion, we filter them with both personal privacy protection and quality check strategies\footnote{Since the Pile is a cleaned high-quality corpus, we directly sample English and code data from the Pile corpus without further filtration. Our filtration strategies are applied to the Chinese data.}, following~\citep{sun2021ernie}.
Our designed filtration strategies includes:
\begin{itemize}
    \item \textbf{Privacy Filtering}: We removed all phone numbers, email addresses, and web links from the corpus to prevent potential privacy breaches.
    \item \textbf{Deduplication}: Basically, we collect our data from different open-sourced datasets with disparate sources. 
    Thus, we mainly conduct deduplication at document, character, and paragraph levels orderly. 
    We treat each sample as a document and use a hash algorithm to remove redundant documents, i.e., retaining only unique documents. 
    Similarly, we also leverage a hash algorithm with an extra sentence segmenter at the paragraph level to identify and remove repeated sentences or paragraphs~(we treat consecutive 1-99 sentences as a paragraph). 
    At the character level, we delete redundant characters and reduce the sequences of repeated characters to a single instance. 
    \item \textbf{Language Filtering}: We use polyglot\footnote{\url{https://github.com/aboSamoor/polyglot}} to detect the language of the text and only keep the texts with high confidence in either Chinese or English. We find it useful to filter out gibberish, especially for texts extracted from PDFs via OCR algorithms.
    \item \textbf{Internet Data Cleaning}: Data collected from the Internet often suffers from incompletions, unrecognizable characters, and web page tags. Thus, we filter out sentences with less than 10 words and remove unusual characters and HTML tags.
\end{itemize}

\begin{figure}[t]
\centering
\subfigure[]{\includegraphics[width=0.32\linewidth]{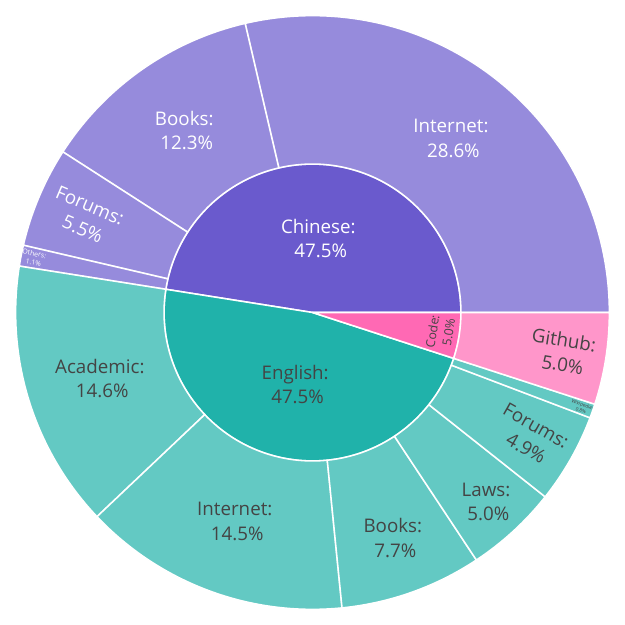}\label{fig:pretrain_collection}}
\hfill
\subfigure[]{\includegraphics[width=0.32\linewidth]{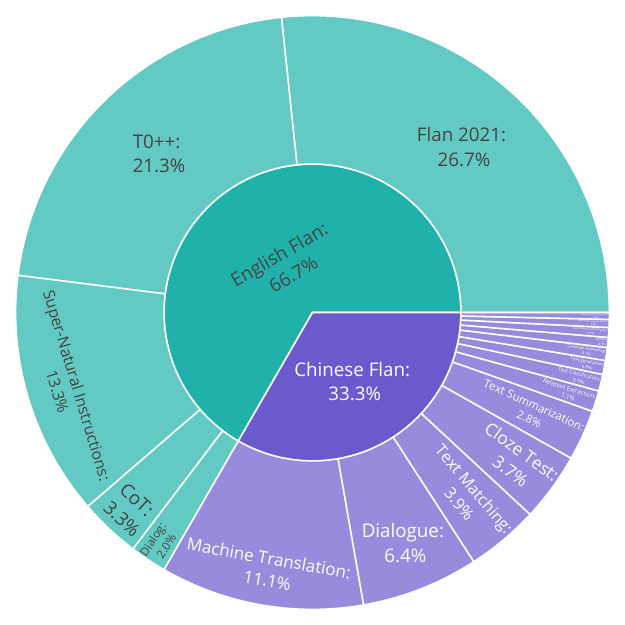}\label{fig:biflan_collection}}
\hfill
\subfigure[]{\includegraphics[width=0.32\linewidth]{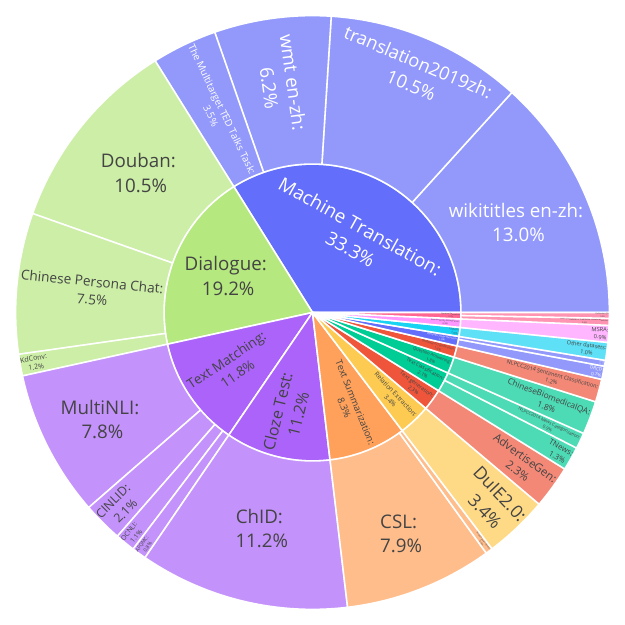}\label{fig:cflan_collection}}
\caption{The composition of Data collection. Figure (a) represents the composition ratio of the pre-training dataset. Figure (b) represents the composition of the Bilingual Flan dataset. Figure (c) represents the finer-grained composition of the Chinese Flan dataset.}
\label{fig:data_collection}
\end{figure}

\paragraph{Mixing and Statistics}
Following~\citep{zeng2022glm}, our pre-training data consists of the same proportion of Chinese and English tokens, in which we sample 190B English tokens\footnote{These English tokens exclude code data but inevitably contain a small percentage of non-language tokens (e.g., 1.24 \% of math data) since we randomly select samples based on the original proportion of the Pile corpus except for code data.} from the Pile corpus and 190B tokens from our filtrated Chinese data.
We also sample 20B code tokens from the Pile corpus to make the overall percentages (5 \%) of code tokens resemble LLaMA~\citep{touvron2023llama} (4.5 \% code tokens).
In total, our pre-training dataset is a mix of 190B English tokens, 190B Chinese tokens, and 20B code tokens.
Following our strategies, one can construct a pre-training dataset with trillion tokens.
Nevertheless, we only collect 400B tokens for pre-training due to our limited computation resources. 
Fig.~\ref{fig:pretrain_collection} shows the detailed composition of the pre-training dataset. 



\subsubsection{Bilingual Flan Data Collection}
\label{subsec:bilingual_flan_data_collection}
\paragraph{English Flan Data Collection}
The Flan Collection~\citep{chung2022scaling, longpre2023flan} serves as a foundational dataset for effective instruction tuning, encompassing more than 1800 tasks. 
We follow the official guidelines to collect and process the English Flan collection with two steps, i.e., downloading five sub-mixtures from the Flan Collection and then combing them according to the specified mixture rates\footnote{\url{https://github.com/google-research/FLAN/tree/main/flan/v2}}.


\paragraph{Chinese Flan Data Collection}
Many open-source Chinese instruction datasets are derived from English translations or generated by ChatGPT using various prompts~\citep{ouyang2022training,bubeck2023sparks}, which can lead to translation inaccuracies and incorrect responses. 
Thus, the quality and quantity of existing Chinese instruction corpora are often inadequate.
To tackle these challenges, we build our own Chinese instruction corpus.
More concretely, we collect 44 different Chinese tasks with a total of 50 million data entries, in which the data sources include various competitions, academic papers, and open-source projects.
The distribution of the Chinese Flan dataset is shown in Fig.~\ref{fig:cflan_collection}.
, and we list all the data sources in Tab.~\ref{tab:chinese_flan}~(Appendix~\ref{appdix:chinese_flan_collection}).
For each task, we manually design the Chinese instructions.

\paragraph{Bilingual Data Combination}
Due to the smaller number of Chinese data compared to English data, we perform sampling within the English Flan datasets to ensure a balanced proportion between Chinese and English data.
As illustrated in Fig.~\ref{fig:biflan_collection}, the Bilingual Flan~(BiFlan) dataset consists of 66.7\% English Flan data and 33.3\% Chinese Flan data. We also filter out samples whose lengths exceed the encoder's maximum length, ensuring the critical parts of instructions are not truncated.


\subsection{Model Architecture}
\label{subsec:model_and_architecture}
Generally, the OpenBA model follows the standard encoder-decoder architecture like T5~\citep{raffel2020exploring}.
However, it is worth noting that the encoder and decoder serve different roles, where the encoder provides strong comprehension capability, while the decoder offers generative ability~\citep{vaswani2017attention}, and existing works indicate that an encoder-decoder model with more encoder layers can achieve powerful performance~\citep{soltan2022alexatm}.
To enhance generative capability and fill the gap of deeper decoder-based LLM, we also design another asymmetric structure, where the detailed model settings are given in Tab.~\ref{tab:model_arch}.
\begin{table}[ht]
    \centering
    \small
    \begin{tabular}{c c c c c c}
    \toprule
     Encoder & Decoder &  Attn Heads & $d_{\mathrm{model}}$ & $d_{\mathrm{ff}}$ & \#Param.(B)\\
     \midrule
     12 & 36 & 40 & 4,096 & 16,384 & 14.6$^{*}$ \\
     \bottomrule
    \end{tabular}
    \caption{Model settings for OpenBA, where \#Param. denotes the number of model parameters. We share the embedding weights between the encoder and decoder, which are not repeatedly counted.}
    \label{tab:model_arch}
\end{table}

Apart from leveraging the asymmetric shallow-encoder deep-decoder, we also incorporate the following improvement strategies into the model architecture:
\begin{itemize}
    \item \textbf{Sandwich Layer Normalization} To stabilize the training process, we adopt the sandwich layer normalization~\citep{ding2021cogview} by normalizing both the input of each transformer block and the output of each attention layer. We use the RMSNorm~\citep{zhang2019root} for normalization. 
    \item \textbf{Rotary Embedding} We apply the rotary embedding~\citep{su2021roformer} that encodes the absolute position with a rotation matrix and meanwhile incorporates the explicit relative position dependency in self-attention instead of the relative positional embedding in T5. 
    \item \textbf{SwiGLU Activation Function} Inspired by LLaMA~\citep{touvron2023llama}, we replace the original ReLU activation with the SwiGLU activation function~\citep{shazeer2020glu}, and set the dimension as $\frac{2}{3}$4$d$. 
    \item \textbf{mT5 Tokenizer} For the bilingual setting, we use mT5~\citep{xue2020mt5} tokenizer as it not only covers Chinese and English but also provides possibilities for further training and applications in other languages.
\end{itemize}

\begin{figure}[t]
    \centering
\includegraphics[width=0.95\textwidth]{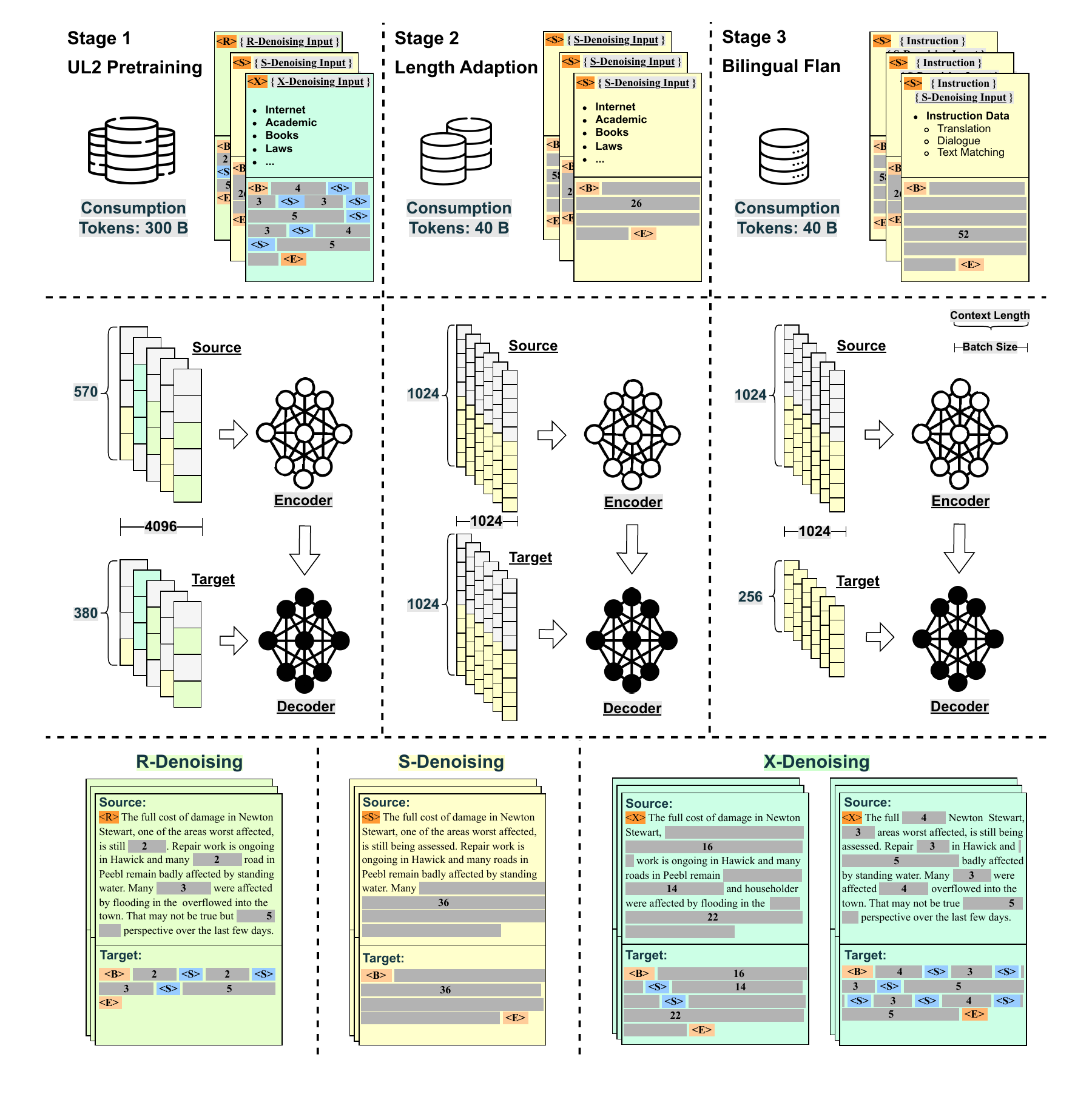}
    \caption{Overview of training process.}
    \label{fig:overview_training_process}
\end{figure}

\subsection{Training Process and Language Modeling Tasks}
\label{subsec:training_process} 
As shown in Fig.~\ref{fig:overview_training_process}, we adopt the stage-wise training strategy~\citep{barshan2015stage} that consists of UL2~\citep{tay2022ul2} pre-training, length-adaptation, and Flan training stages~\citep{wei2021finetuned}, and set different context length and batch size for different stages~(Tab.~\ref{tab:context_length}).
\begin{table}[t]
    \centering
    \small
    \resizebox{\textwidth}{!}{
    \begin{tabular}{c c c c c c}
    \toprule
    \bf Stage & \bf Strategy & \bf Encoder Context & \bf Decoder Context & \bf Batch Size & \bf \#Tokens~(B) \\
    \midrule
    \uppercase\expandafter{\romannumeral1} & UL2 Pre-training & 570 & 380 & 4,096 & 300 \\
    \uppercase\expandafter{\romannumeral2} & Length-Adaptation & 1,024 & 1,024 & 1,024 & 40 \\
    \uppercase\expandafter{\romannumeral3} & Bilingual Flan Training & 1,024 & 256 & 1,024 & 40 \\
    \bottomrule
    \end{tabular}}
    \caption{Configurations for each training stage, where \#Tokens represents the number of consumption tokens in each stage.}
    \label{tab:context_length}
\end{table}
For all the stages, we adopt the span-denoising language modeling task as proposed in T5~\citep{raffel2020exploring} and BART~\citep{lewis2019bart}.
More concretely, given a sequence $\boldsymbol{x}=\{x_{i}\}_{i=1}^{n}$ containing $n$ tokens, we corrupt it with mask sentinel $m_{j}=\{x_{i}\}_{i=p}^{k}$, where $p < k, 1 \leq p, k \leq n$.
Then, the training objective is to recover the masked span, which can be formally written as: 
\begin{equation}
    \mathcal{L}(\boldsymbol{x}) = \log P(\boldsymbol{m} \mid \boldsymbol{x}_{\backslash \boldsymbol{m}}, \theta)
\end{equation}
where $\boldsymbol{m} = \{m_{j}\}_{j=1}^{K}$ is the set of masked spans, $\boldsymbol{x}_{\backslash \boldsymbol{m}}$ denotes the corrupted sequence, and $\theta$ represents the model parameters. 
For OpenBA model, $\boldsymbol{x}_{\backslash \boldsymbol{m}}$ is fed to the encoder, which can retain a bidirectional receptive field, and $\boldsymbol{m}$ is predicted by the decoder.
Next, we will introduce the aforementioned different stages more concretely.

\paragraph{Stage~\uppercase\expandafter{\romannumeral1}: UL2 Pre-training}
Starting with the UL2 training strategy, we adopt a mixture of denoisers training strategy proposed by~\citet{tay2022ul2}, exposing OpenBA to a diverse set of problems during the first pre-training stage.
In this stage, the OpenBA model is fed with 300B tokens sampled from the pre-training corpus, and we employ three different training objectives which are listed below: 
\begin{itemize}
\item \textbf{R-Denoising} Regular denoising is the standard span corruption that sets a range of 2 to 5 tokens as the masked span length and masks ratio about 15\% of the input tokens. This denoising task is relatively simple since the span is short and efficient for the model to acquire knowledge embedded in the text.
\item \textbf{S-Denoising} Sequence denoising aims to endow the model with generation capability, where the input text is split into two sub-sequences, and the model should predict the latter sequence conditioned on the first sequence. In the S-Denoising setting, the model can acquire the generation ability.
\item \textbf{X-Denoising} To bridge the gap between the R-Denoising and S-Denoising, X-Denoising can be viewed as an extreme version of denoising, where approximately 50\% of the input sequence is masked by increasing either the masked span length or the corruption rate. 
Such a denoising strategy simulates the situation where a model needs to generate long targets from a memory with relatively limited information.

\end{itemize}

We list the settings of these three denoising strategies in Tab.~\ref{tab:ul2_denoiser_setting}. 
It is worth noting that we conduct these denoising strategies from the instance level and prepend three special tokens before each corrupted sequence to prompt the current denoising task for OpenBA~\citep{tay2022ul2}.
We uniformly sample a value based on $\mu$ as the masked span length for the R-denoising and X-denoising.
For S-denoising, we limit each masked span to end at the end of the input text and allow only one masked span.
Besides, we set encoder-decoder context length as 570/380 in this stage for sampling efficiency.

\begin{table}[t]
    \centering
    \small
    \begin{tabular}{l c  c  c c}
    \toprule
    \bf Type & \bf Span Length ($\mu$) & \bf Corruption Ratio (\%) & \bf \#Num & \bf Sentinel \\
    \midrule
     R-Denoising &  \{3, 8\} & 15.0 & $K$ & \texttt{<R>}\\
     S-Denoising & - & 25.0 & 1 & \texttt{<S>} \\
     X-Denoising & \{3, 8, 64\} / \{64\} & 50.0 / 15.0 & $K$ & \texttt{<X>} \\
    \bottomrule
    \end{tabular}
    \caption{Settings of three denoising strategies for the UL2 pre-training stage, where $\mu$ is the mean of the normal distribution, \#Num represents the number of masked spans, and $K$ is determined by the sequence length, span length, and corruption ratio.}
    \label{tab:ul2_denoiser_setting}
\end{table}

\paragraph{Stage~\uppercase\expandafter{\romannumeral2}: Length-Adaptation}
Considering the context length for the first pre-training stage is short, which may not support the long input and output formats of some tasks, such as in-context learning~\citep{min2021metaicl} and long text generation~\citep{guan2021long}, we extend the encoder-decoder context length to 1,024/1,024 during the length-adaptation stage.
During this stage, we utilize 40B tokens sampled from the pre-training corpus and ensure that there is no overlap between these data and the data from the previous stage.
Additionally, we simply apply the S-Denoising training objective and adjust the corruption ratio to 50\%.
We keep the special sentinel $\texttt{<S>}$ before each corrupted text and decrease the batch size for training stability in this stage.

\paragraph{Stage~\uppercase\expandafter{\romannumeral3}: Bilingual Flan Training}
Inspired by the previous work~\citep{chung2022scaling}, we apply Flan instruction training on the length-adapted OpenBA checkpoint.
We still prepend the special token $\texttt{<S>}$ before each text for the generation task and apply the constructed BiFlan dataset in this stage.
In addition, we set the encoder-decoder sequence length as 1,024/256 in this stage for sampling efficiency since we observe that most outputs of Flan datasets are short, i.e., less than 256 tokens.

\begin{figure}[t]
\centering
\subfigure[Training loss of Stage \uppercase\expandafter{\romannumeral1}]{\includegraphics[width=0.9\linewidth]{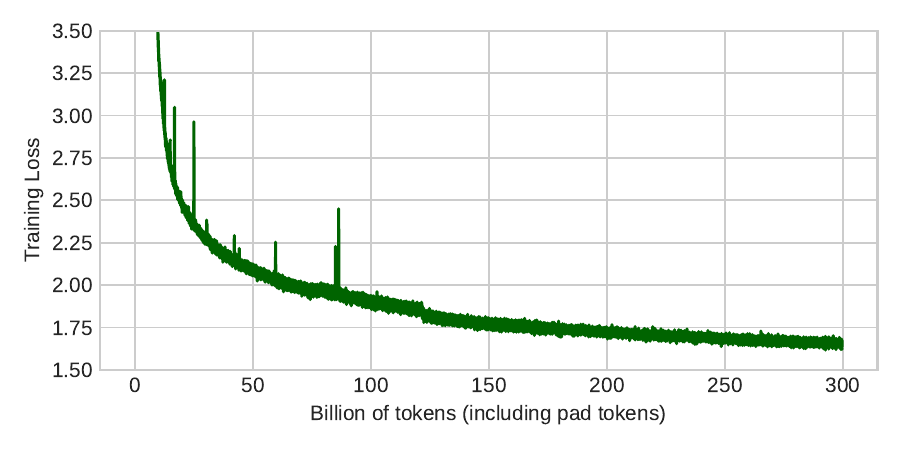}
\label{fig:losspretrain}} \\
\subfigure[Training loss of Stage \uppercase\expandafter{\romannumeral2}]{\includegraphics[width=0.45\linewidth]{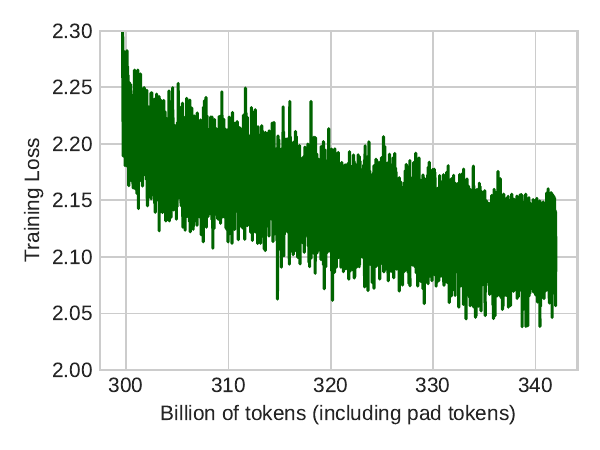}\label{fig:lossstretch}}
\subfigure[Training loss of Stage \uppercase\expandafter{\romannumeral3}]{\includegraphics[width=0.45\linewidth]{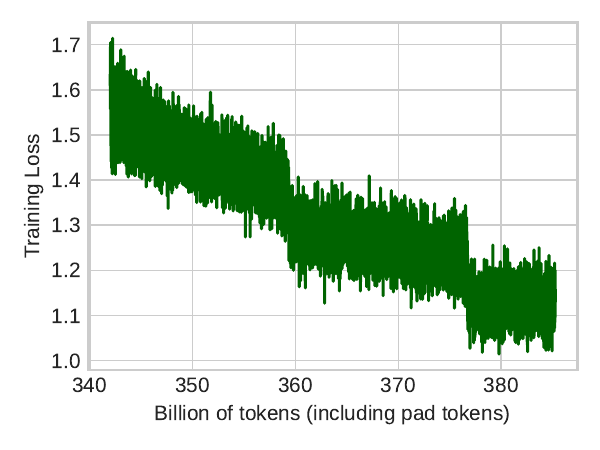}\label{fig:lossflan}}
\caption{Loss curves for each training stage.}
\label{fig:lossacctask}
\end{figure}

\subsection{Model Implementation and Techniques}
\label{subsec:model_implementation}
We train OpenBA on a cluster with 4 nodes (8 $\times$ NVIDIA A100-SXM4-80GB GPUs), which are linked with the InfiniBand network~\citep{grun2010introduction} and interconnected through the NVLink system. 
The model has consumed nearly 400B bilingual tokens and achieved $1.2\times 10^{22}$ FLOPs~(floating point of operations) in total.
We implement our model based on the NVIDIA-Megatron framework\footnote{\url{https://github.com/NVIDIA/Megatron-LM/}} and make several optimizations for training stabilization and inference efficiency.
We plot the training loss for the aforementioned three stages in Fig.~\ref{fig:lossacctask}, and list the techniques we 
 have used below:

\paragraph{$\bullet$ 3D Parallelism}
3D parallelism~\citep{shoeybi2019megatron} aims to scale and accelerate the training process of LLMs, which harnesses three core parallelism techniques, i.e., data parallelism, model parallelism (mp), and pipeline parallelism (pp). Considering the model size, the number of GPUs and the communication speed among GPUs, we settle on an optimal setting of \textit{mp\_size}=4 and \textit{pp\_size}=1, reaching 120 TFLOP/s per GPU.

\paragraph{$\bullet$ Checkpoint Activation}

Checkpoint activation is a technique designed to optimize memory usage during training. 
Instead of storing all intermediate layer activations, only certain ones are preserved.
During back-propagation, the missing activations are recalculated, trading off additional computational efforts for memory savings.
This strategy allows for the training of larger models even on GPUs with limited memory capacity.
In fact, training a 15B model on 80GB GPUs becomes manageable in terms of memory.
We specifically apply the checkpoint activation to the attention computation, which is relatively cost-effective to recompute.
In practical deployment, we observe a significant improvement in GPU memory utilization, enhancing the overall system performance.

\paragraph{$\bullet$ Distributed Optimizer}
The distributed optimization approach offers an alternative for saving GPU memory, enabling the utilization of an increased batch size, albeit at the expense of communication burden among GPUs.
By adopting the ZeRO method proposed by \citet{rajbhandari2020zero} and implementing the distributed optimization technique~\citep{shoeybi2019megatron}, we can increase the batch size, thereby enhancing the training speed.

\paragraph{$\bullet$ Attention Weights Computation in FP32 Precision}
During the softmax computation, particularly when handling large values, there exists a possibility of numerical overflow.
Conducting this computation with FP32 precision mitigates this risk compared to using FP16 precision.
The previous works~\citep{nijkamp2022codegen} indicate that such an issue can easily take place when computing attention weights in FP16 precision.
In the early training stage of OpenBA, we adopt half-precision calculation for all the model modules and often observe the phenomenon of loss collapsing. 
However, such an issue has been greatly alleviated when converting the attention weight calculations to full precision~(FP32).
Thus, we can empirically conclude that attention weight computation in FP32 precision can significantly enhance the stability of the training process.

\paragraph{$\bullet$ Inference Efficiency} 
To accelerate the inference speed, we adopt the KV-cache technique and decrease the computation by pre-computing the rotary embeddings for all the positions.


\section{Results}
\label{sec:results}


\begin{table}[t]
    \centering
    \small
    \resizebox{\textwidth}{!}{
    \begin{tabular}{l c c c c c c}
    \toprule
     \bf Model & \bf \#Param. & \bf Tokens & \bf GPU/TPU type & \bf GPU hours & \bf \makecell[c]{Total Power \\ Consumption} & \bf \makecell[c]{Carbon emitted \\ (tCO$_{2}$eq)} \\
     \midrule
     OPT~\cite{zhang2022opt} & 175B & 180B & A100-80GB & 809,472 & 356 MWh & 137 \\
     BLOOM~\cite{scao2022bloom} & 176B & 366B & A100-80GB & 1,082,880 & 475 MWh & 183 \\
     GLM~\cite{zeng2022glm} & 130B & 400B & A100-40GB & 1,105,920 & 442 MWh & 257 \\
     ChatGLM~\cite{zeng2022glm} & 6B & 1.0T & - & - & - & - \\
     Falcon~\cite{refinedweb} & 40B & 1.0T & A100-40GB & - & - & - \\
     Flan-T5-XL~\cite{chung2022scaling} & 3B & >1.0T & TPU-v3/v4 & - & - & - \\
     LLaMA~\cite{touvron2023llama} & 7B & 1.0T & A100-80GB & 82,432 & 36 MWh & 14\\
     LLaMA~\cite{touvron2023llama} & 13B & 1.0T & A100-80GB & 135,168 & 59 MWh & 23 \\
     LLaMA~\cite{touvron2023llama} & 65B & 1.4T & A100-80GB & 1,022,362 & 449 MWh & 173 \\
     LLaMA-2-Chat~\cite{touvron2023llama2} & 70B & >2.0T & A100-80GB & 1,720,320 & - & 291 \\
     Baichuan~\cite{Baichuan2023} & 7B & 1.2T & A800 & - & - & - \\
     BatGPT~\cite{li2023batgpt} & 15B & 1.0T & - & - & - & - \\
     MOSS~\cite{sun2023moss} & 16B & >700B & - & - & - & - \\
     \midrule
     OpenBA & 15B & 380B & A100-80GB & 38,214 & 17 MWh & 6.5 \\
     \bottomrule
    \end{tabular}}
    \caption{The number of parameters, consumed tokens, and training cost for the LLMs mentioned in the paper, where \#Param. denotes the model parameters. We report the carbon emission according to the official statement, and calculate the carbon emission of OpenBA according to~\citet{wu2022sustainable}.
    }
    \label{tab:carbon_footprint}
\end{table}

\subsection{Evaluation Settings}
\label{subsec:experimantal_settings}
We evaluate OpenBA from three aspects: natural language understanding, natural language generation, and commonsense reasoning.
Specifically, we evaluate the natural language understanding capability on the SuperGLUE~\citep{wang2019superglue}  and BELEBELE~\citep{bandarkar2023belebele} benchmark, natural language generation ability with five downstream tasks (summarization, machine translation, text simplification, paraphrase, and story generation), and commonsense reasoning ability on five authoritative benchmarks, including MMLU~\citep{hendrycks2020measuring}, CMMLU~\citep{li2023cmmlu}, BBH~\citep{suzgun2022challenging}, and C-Eval~\citep{huang2023c}.
Following the previous works~\citep{brown2020language,touvron2023llama}, we consider both the zero-shot and few-shot settings and strictly distinguish the domain distribution of training and testing data.
The illustration and the corresponding implementation of each setting are as follows:
\begin{itemize}
    \item \textbf{Zero-Shot} We provide a textual description of the task for each testing sample, and the model will respond in an open-ended manner. Templates for all tasks are listed in Appendix~\ref{sec:instruction_template}.
    \item \textbf{Few-Shot} We evaluate each example in the testing set by randomly selecting $\ell$ examples from the training set of each task as conditioning. In this paper, we set $\ell=5$ as default if not specified.
    \item \textbf{Domain Held-in / Held-out} We differentiate between the held-in and held-out settings based on whether the training data includes the domain of the testing set. If the model has been trained on the training data corresponding to the testing task, it is viewed as held-in; otherwise, it is held-out~\citep {longpre2023flan}.
\end{itemize}

We also apply the CoT technique for some tasks, and the corresponding templates are also shown in Appendix~\ref{sec:instruction_template}.
It is worth noting that we will specifically elaborate on the basic settings for each evaluation task and compare them to the models under the same settings. 
Additionally, we will evaluate the results using the officially recommended evaluation metrics and platforms whenever possible and utilize the \textbf{\textit{bold font}} to indicate \textbf{\textit{the best performance}} and adopt \underline{\textit{underline}} to denote \underline{\textit{the second-best}} performance in all the experiments.

\subsection{Training Cost Analysis}
\label{subsec:llm_selection}
All the models we compare are listed in Tab.~\ref{tab:carbon_footprint}, where we report their parameters, consumption tokens, training cost, and the corresponding carbon emissions, respectively.
To calculate carbon emissions, we follow~\citet{wu2022sustainable} and~\citet{touvron2023llama} by taking a PUE of 1.1 and a carbon intensity factor set at the national US average of 0.385 kg CO2e per KWh, and the formula is:
\begin{equation}
    tCO_{2eq} = MWh \times 0.385
\end{equation}
It is worth noting that, \textbf{\textit{the training process of OpenBA is highly efficient and environmentally friendly.}}
Taking \textbf{\textit{LLaMA-13B}} as an example, it consumes around 1TB tokens with a total 59MWh GPU power and emits around \textbf{\textit{23 $tCO_{2eq}$ carbon}}.
However, our model has consumed only 6.5 $tCO_{2eq}$ carbon for 380B tokens, i.e., around \textbf{\textit{28.26 \% of the total carbon emission of the LLaMA-13B model}}.
More training details and model implementation can be found in Sec.~\ref{subsec:model_implementation}.

\subsection{Natural Language Understanding}
\begin{table}[t]
    \centering
    \small
    \begin{tabular}{l c c c c c c c c}
    \toprule
    \bf Model & \bf \#Param. & \bf Avg. & \bf BoolQ & \bf CB & \bf RTE & \bf ReCoRD & \bf ReCoRD & \bf WSC \\
    \bf Metrics & & & \bf Acc. & \bf Acc. & \bf Acc.& \bf F1 & \bf EM & \bf Acc. \\
    \midrule
    BERT-Large & 340M & 69.0 & 77.4 & 83.6 & 71.6 & \underline{72.0} & 71.3 & 64.3 \\ 
    BERT-Large++ & 340M & 71.5 & 79.0 & \underline{90.4} & 79.0 & \underline{72.0} & \underline{73.0} & 64.3 \\
    Flan-T5-XL & 3B & \bf 79.3 & \bf 89.3 & \bf 91.2 & \bf 90.4 & 57.2 & 56.6 & \bf 84.9 \\
    GPT3 & 175B & 71.8 & 76.4 & 75.6 & 69.0 & \bf 91.1 & \bf 90.0 & \underline{80.1} \\
    \midrule
    OpenBA & 15B & \underline{73.1} & \underline{82.6} & 85.6 & \underline{83.9} & 69.4 & 68.8 & 76.0 \\
    \midrule
    \bf Model & \bf \#Param. & \bf WiC  &\bf CoPA & \bf MultiRC & \bf MultiRC & \bf AX$_b$ & \bf AX$_g$ & \bf AX$_g$ \\
    \bf Metrics & &\bf Acc. & \bf Acc. & \bf F1 & \bf EM & \bf MCC & \bf GPS & \bf Acc \\
    \midrule
    BERT-Large & 340M & \bf 69.5 & 70.6 & 70.0 & 24.0 & 23.0 & \underline{97.8} & 51.7 \\
    BERT-Large++ & 340M & \bf 69.5 & 73.8 & 70.4 & 24.5 & 38.0 & \bf 99.4 & 51.4 \\
    Flan-T5-XL & 3B & 65.7 & \bf 97.6 & \bf 87.0 & \bf 57.9 &  \bf 50.1 & 97.2 & \bf 91.9 \\
    GPT3 & 175B & 49.4 & \underline{92.0} & 75.4 & 30.5 &  21.1 & 90.4 & 55.3 \\
    \midrule
    OpenBA & 15B & 57.2 & 85.8 & \underline{77.1} & \underline{38.9} &\underline{40.8} & 94.4 & \underline{70.2} \\
    \bottomrule
    \end{tabular}
    \caption{Zero-shot results on SuperGLUE benchmark, where \#Param. denotes the model parameters, and Avg. denotes average accuracy.}
    \label{tab:superglue_results}
\end{table}

We evaluate the natural language understanding performance of OpenBA model on the SuperGLUE benchmark, which contains 13 sub-tasks. 
Since the BiFlan dataset contains partial training data of some testing tasks in SuperGLUE, we mainly compare OpenBA with models in the held-in setting (except GPT-3~\citep{brown2020language}), i.e., these models have also been trained on the training data of some testing tasks in SuperGLUE.
As we can observe in Tab.~\ref{tab:superglue_results}, the performance of OpenBA surpasses that of the BERT model~\citep{devlin2018bert} fine-tuned on the SuperGLUE training set and GPT-3, but is slightly behind that of the Flan-T5-XL~\citep{chung2022scaling} model.

\begin{table}[t]
    \centering
    \small
    \begin{tabular}{l c c c c c c }
    \toprule
    \bf Model & \bf \#Param. & \bf eng\_Latn & \bf zho\_Hans & \bf zho\_Hant & \bf Avg. \\
    \midrule
    Falcon$^{\dagger}$ & 40B & 77.2 & 66.0 & 62.2 & 68.5 \\
    LLaMA$^{\dagger}$ & 70B  & \underline{82.5} & 64.6 & 57.7 & 68.2 \\
    InfoXLM$^{\ddagger}$ & 550M & 79.3 & 74.6 & 72.4 & 75.4 \\ 
    XLM-V$^{\ddagger}$ & 1.2B & 76.2 & 71.0 & 67.1 & 71.4 \\ 
    LLaMA-2-Chat$^{*}$ & 70B & 78.8 & 62.4 & 59.3 & 66.8 \\
    GPT3.5-Turbo$^{*}$ & - & \bf 87.7 & \bf 77.6 & \bf 76.3 & \bf 80.5 \\
    \midrule
    OpenBA$^{*}$ & 15B  & 78.6 & \underline{75.2} & \underline{73.7} & \underline{75.8} \\
    \bottomrule
    \end{tabular}
    \caption{Model performance on BELEBELE benchmark, where $\dagger$ denotes 5-shot setting, $\ddagger$ denotes full fine-tuning in English and $*$ denotes the zero-shot setting for instructed models. We report the accuracy score for all the models.}
    \label{tab:belebele_results}
\end{table}

We evaluate the reading comprehension ability of OpenBA with BELEBELE benchmark~\citep {bandarkar2023belebele} and select the Chinese~(Simplified), Chinese~(Traditional), and English subsets for evaluation.
We follow the official settings and compare with both LLMs and fine-tuned down-stream models, including Falcon~\citep{penedo2023refinedweb}, LLaMA~\citep{touvron2023llama,touvron2023llama2}, XLM-V~\citep{liang2023xlm}, InfoXLM~\citep{chi2020infoxlm} and ChatGPT~\citep{ouyang2022training}. 
We provide all the instructions we use for zero-shot setting in Appendix~\ref{sec:instruction_template}.
As we can observe from Tab.~\ref{tab:belebele_results}, OpenBA can achieve outstanding results in the Chinese reading comprehension tasks, ranking just behind ChatGPT.
For English reading comprehension tasks, the performance of OpenBA is comparable to that of the Falcon-40B model, which is trained with around 1TB tokens of multilingual data.
It is also worth noting that OpenBA achieves better performance among multiple current open-source LLMs, including two strong LLaMA models and the Falcon-40B model, under the bilingual setting.

\subsection{Natural Language Generation}
We evaluate the natural language generation ability of our model on five tasks, including machine translation on the Flores~\citep{goyal2022flores} benchmark, text summarization on the CLTS benchmark~\citep{liu2020clts}, paraphrase task on the QQP dataset\footnote{\url{https://www.kaggle.com/c/quora-question-pairs}}, text simplification on the WIKI-AUTO~\citep{coster2011simple} dataset, and story generation on the ROC~\citep{mostafazadeh2016corpus} dataset.

\paragraph{Summarization}

\begin{table}[t]

    \begin{minipage}{0.5\textwidth}
    \centering
    \small
    \resizebox{\textwidth}{!}{
    \begin{tabular}{>{\rule{0pt}{2.5ex}}l >{\rule{0pt}{2.5ex}}c >{\rule{0pt}{2.5ex}}c >{\rule{0pt}{2.5ex}}c >{\rule{0pt}{2.5ex}}c}
    \toprule
    \bf Model & \bf \#Param. & \bf Rouge-1 & \bf Rouge-2 & \bf Rouge-L \\
    \midrule
    ChatGLM & 6B & \underline{27.3} & \textbf{17.2} & \underline{26.7} \\
    Baichuan & 7B & 19.9 & \underline{14.4} & 20.0 \\
    BatGPT & 15B & 25.6 & 12.2 & 25.0 \\
    OpenBA & 15B & \textbf{30.2} & 13.9 & \textbf{28.6} \\
    \bottomrule
    \end{tabular}}
    \caption{Model performance on CLTS subset containing 100 sentences sampled from CLTS test set. We report Rouge-1, Rouge-2 and Rouge-L score.}
    \label{tab:summarization_res}
    \end{minipage}
    \hfill
    \begin{minipage}{0.45\textwidth}
        \small
        \centering
        \resizebox{\textwidth}{!}{
        \begin{tabular}{l c c c}
            \toprule
            \bf Model & \bf \#Param. & \textbf{Zh}$\Rightarrow$\textbf{En} & \textbf{En}$\Rightarrow$\textbf{Zh} \\
            \midrule
            ChatGLM & 6B & 17.2 & 32.5 \\
            Alpaca & 7B & 15.1 & 9.8 \\
            PARROT & 7B & 19.6 & 24.8 \\
            BatGPT & 15B & \underline{23.1} & \bf 38.7 \\
            MOSS & 16B & 17.2 & 32.5 \\
            \midrule
            OpenBA & 15B & \textbf{23.3} & \underline{37.4} \\
            \bottomrule
        \end{tabular}}
        \caption{Model performance on Flores subset containing 50 sentences sampled from Flores.}
        \label{tab:mt_results}
    \end{minipage}
\end{table}
To evaluate the summarization task under the held-out setting, we select a subset containing 100 sentences sampled from CLTS benchmark~\citep{liu2020clts}, which is excluded from the BiFlan dataset.
Specifically, we prepend the task instruction before each test sentence (the task instruction is listed in Appendix~\ref{sec:instruction_template}) and allow models to conduct zero-shot inference.
We evaluate the generated results with Rouge-$n$ metric~\citep{lin2004rouge} and report the results in Tab.~\ref{tab:summarization_res}.
We observe that OpenBA can achieve the best performance on the Rouge-1 and Rouge-L scores, indicating that the content generated from OpenBA is faithful to the original text in the summarization task.

\begin{table}[t]
    \centering
    \small
    \resizebox{\textwidth}{!}{
    \begin{tabular}{l c | c c c c c | c c c c c}
    \toprule
    \multirow{2}{*}{\bf Model} & \bf \multirow{2}{*}{\#Param.} & \multicolumn{5}{c|}{\bf WIKI AUTO} & \multicolumn{5}{c}{\bf QQP}\\
    \cmidrule{3-12}
    & & \bf B-2~($\uparrow$) & \bf D-2~($\uparrow$) & \bf LR-2~($\downarrow$) & \bf Mav~($\uparrow$) & \bf SIM~($\uparrow$)& \bf B-2~($\uparrow$) & \bf D-2~($\uparrow$) & \bf LR-2~($\downarrow$) & \bf Mav~($\uparrow$) & \bf SIM~($\uparrow$)  \\
    \midrule
    BatGPT & 15B & 25.5 & \underline{1.5} & 89.9 & 96.5 & \underline{5.5} & 19.4 & 1.6 & 67.6 & 58.3 & 7.6 \\
    chatGLM & 6B & \textbf{29.2} & 1.4 & \underline{90.7} & \underline{97.7} & 4.0 & \textbf{25.0} & \underline{2.0} & 63.9 & \underline{93.6} & 5.6 \\
    MOSS & 16B & 27.8 & \underline{1.5} & 82.9 & 96.8 & 5.4 & 19.3 & 1.4 & \underline{72.7} & 37.2 & \underline{7.8} \\
    OpenBA & 15B & \underline{27.9} & \textbf{1.9} & \textbf{75.6} & \textbf{99.1} & \textbf{6.6} & \underline{22.7} & \textbf{2.0} & \textbf{48.0} & \textbf{94.4} & \textbf{7.9} \\
    \bottomrule
    \end{tabular}}
    \caption{Model performance on WIKU AUTO and QQP datasets.}
    \label{tab:wiki_qqp}
\end{table}

\begin{figure}[t]
\centering
\subfigure[Coherence]{\includegraphics[width=0.8\linewidth]{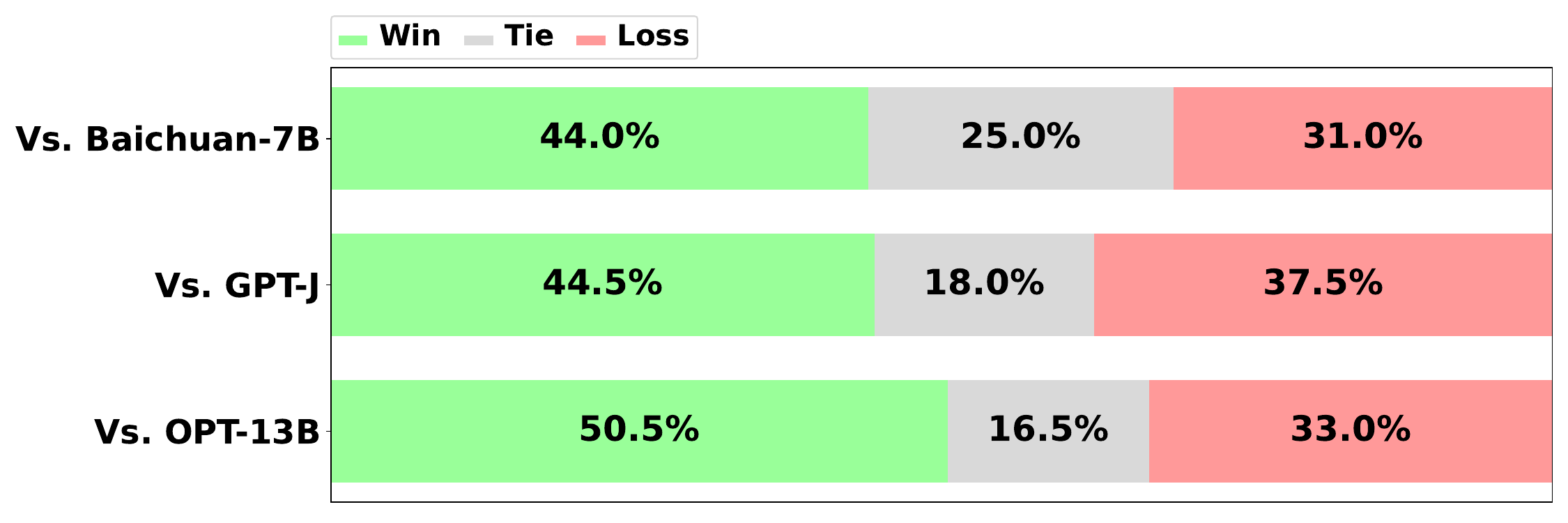}\label{fig:humaneval_coherence}} \\
\subfigure[Consistency]{\includegraphics[width=0.8\linewidth]{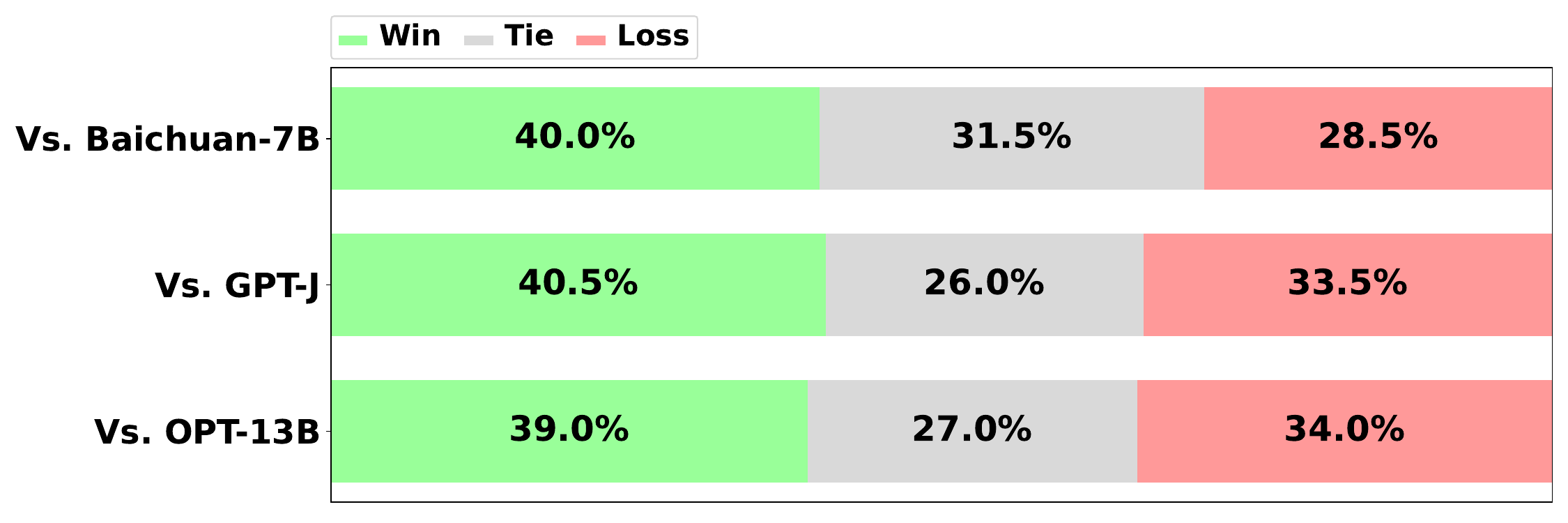}\label{fig:humaneval_consistency}} \\
\subfigure[Correctness]{\includegraphics[width=0.8\linewidth]{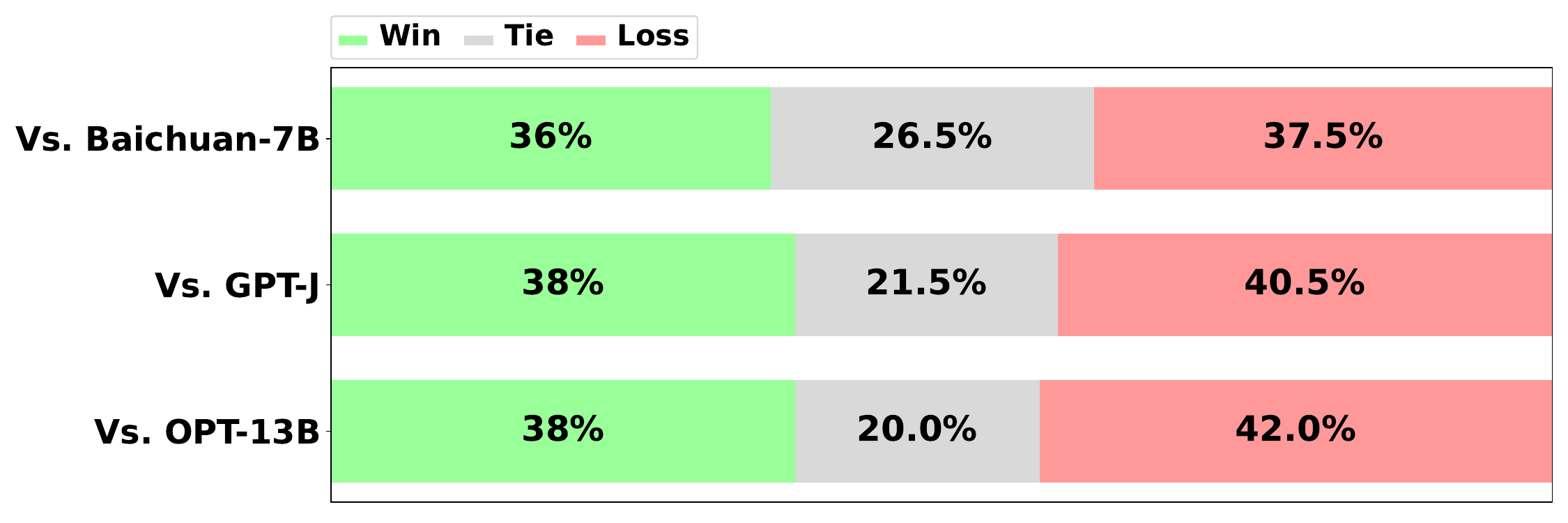}\label{fig:humaneval_correctness}} \\
\caption{Human evaluation results on the ROC dataset.}
\label{fig:humanevalres}
\end{figure}

\paragraph{Machine Translation} 
We compare the model performance on the bilingual machine translation tasks, including Chinese-to-English and English-to-Chinese translation, on the Flores~\citep{goyal2022flores} machine translation benchmark.
We strictly follow the official settings by selecting 50 testing samples provided for each translation task.
It is worth noting that all the models are under the held-out zero-shot setting.
We report the BLUE-4~\citep{post-2018-call} scores in Tab.~\ref{tab:mt_results} and can observe that OpenBA can achieve the best performance on the Chinese-to-English translation task and obtain comparable results with the SOTA achieved by BatGPT on the English-to-Chinese translation task.

\paragraph{Text Simplification and Paraphrase}
We evaluate the text simplification and paraphrase ability of OpenBA on the WIKI AUTO and QQP datasets.
We evaluate the model performance with BLUE, Distinct-$n$~(D-$n$) metrics~\citep{li2015diversity}, Lexical Repetition (Rep-$n$, 4-gram repetition for $n$-times)~\citep{shao2019long}, Mauve~\citep{pillutla2021mauve} and Semantic Similarity (SIM, semantic similarity between generations and corresponding
prompts)~\citep{guan2021long} metrics, and report the model performance in Tab.~\ref{tab:wiki_qqp}.
Based on the observation that OpenBA can attain the best results on the Mav and SIM metrics, 
 which evaluate semantic relevance with gold text and input text respectively, we can conclude that our model excels at capturing the overall semantic information of the input content and generating relevant content accordingly.

\begin{table}[t]
    \centering
    \small
    \begin{tabular}{l c c c c c c}
    \toprule
    \bf Model & \bf \#Param. & \bf Humanities & \bf STEM & \bf Social Sciences & \bf Other & \bf Average \\
    \midrule
    LLaMA$^{\dagger}$ & 7B & 34.0 & 30.5 & 38.3 & 38.1 & 35.1 \\
    LLaMA$^{\dagger}$ & 13B & \textbf{45.0} & \underline{35.8} & \textbf{53.8} & \textbf{53.3} & \textbf{46.9} \\
    BLOOM$^{\dagger}$ & 176B & 34.1 & \textbf{36.8} & 41.5 & 46.5 & 39.1 \\
    \midrule
    ChatGLM$^{\dagger}$ & 6B & 35.4 & 31.3 & 41.0 & 40.5 & 36.9 \\
    Baichuan$^{\dagger}$ & 7B & 38.4 & 35.6 & \underline{48.9} & \underline{48.1} & \underline{42.3} \\
    BatGPT$^{\dagger}$ & 15B & 35.4 & 33.5 & 36.3 & 37.0 & 36.7 \\
    MOSS$^{\dagger}$ & 16B & 30.5 & 29.3 & 33.8 & 34.4 & 31.9 \\   
    \midrule
    OpenBA$^{\dagger}$ & 15B & 34.6 & 29.8 & 40.1 & 40.0 & 36.0 \\
    OpenBA$^{\ddagger}$ & 15B & \underline{38.7} & 33.8 & 45.0 & 43.6 & 40.2 \\
    OpenBA$^{*}$ & 15B & 36.7 & 31.4 & 42.8 & 42.3 & 38.2 \\ 
    \bottomrule
    \end{tabular}
    \caption{Model performance on MMLU benchmark, where \#Param. denotes the model parameters, $\dagger$ denotes 5-shot, $\ddagger$ denotes 0-shot, and $*$ represents the chain-of-thought. }
    \label{tab:mmlu_results}
\end{table}
\begin{table}[t]
    \scriptsize
    \centering
    \resizebox{\textwidth}{!}{
    \begin{tabular}{l c c c c c c c}
    \toprule
    \bf Model & \bf \#Param. & \bf Humanities & \bf STEM & \bf Social Science & \bf Other & \bf China-specific & \bf Average \\
    \midrule
    Falcon & 40B & \underline{43.5} / 41.3 & 33.3 / 31.1 & 44.3 / 40.9 & \underline{44.8} / 40.6 & \underline{39.5} / 36.1 & \underline{41.5} / 38.5 \\
    LLaMA & 65B & 40.2 / 34.5 & 34.5 / 31.1	& 41.6 / 36.1 & 42.9 / 37.9 & 37.0 / 32.9 & 39.8 / 34.9 \\
    \midrule
    ChatGLM & 6B & 39.2 / 42.9 & 32.4 / 32.2 & 39.7 / 44.8 & 38.6 / 42.6 & 37.7 / 41.9 & 37.5 / 40.8 \\
    Baichuan & 7B & \textbf{48.1} / 44.4 & \textbf{35.3} / 32.8 & \textbf{47.9} / 46.8 & \textbf{46.6} / 44.8 & \textbf{44.1} / 43.1 & \textbf{44.4} / 42.3 \\
    BatGPT & 15B & 35.5 / 36.5 & \underline{35.0} / 33.7 & 36.3 / 38.1 & 42.1 / 46.9 & 37.9 / 38.3 & 37.2 / 38.5 \\
    \midrule
    OpenBA & 15B & 40.9 / 40.9 & 33.5 / 33.8 & \underline{45.2} / 44.7 & 44.5 / 43.6 & 39.1 / 38.6 & \underline{41.5} / 41.2 \\
    OpenBA$^{*}$ & 15B & 30.0 & 37.6 & 40.6 & 39.2 & 36.4 & 37.0 \\
    \bottomrule
    \end{tabular}}
    \caption{Performance on CMMLU benchmark, where \#Param. denotes the model parameters, and $*$ denotes chain-of-thought. We report the 5-shot and 0-shot performance with diagonal bar division.}
    \label{tab:cmmlu_results}
\end{table}

\begin{table}[t]
    \centering
    \small
    \resizebox{\textwidth}{!}{
    \begin{tabular}{l c c c c c c c}
    \toprule
    \bf Model & \bf \#Param. & \bf STEM & \bf Social Science & \bf Humanities & \bf Others & \bf Avg. & \bf Avg.(Hard) \\
    \midrule
    LLaMA & 65B  & \underline{37.8} & 45.6 & 36.1 & 37.1 & 38.8 & \underline{31.7} \\
    \midrule
    ChatGLM & 6B   & 33.3 & 48.3 & 41.3 & 38.0 & 38.9 & 29.2 \\
    Baichuan & 7B   & \bf 38.2 & \underline{52.0} & \underline{46.2} & 39.3 & \underline{42.8} & \bf 31.5 \\
    MOSS-moon-sft & 16B  & 31.6 & 37.0 & 33.4 & 32.1 & 33.1 & 28.4 \\
    GLM-130B & 130B & 36.7 & \bf 55.8 & \bf 47.7 & \bf 43.0 & \bf 44.0 & 30.7 \\
    \midrule
    OpenBA & 15B & 34.8 & 46.6 & 41.1 & \underline{41.5} & 39.8 & 31.1 \\
    OpenBA$^{*}$ & 15B & 30.7 & 43.7 & 40.9 & 35.2 & 36.3 & 27.0 \\
    \bottomrule
    \end{tabular}}
    \caption{Model performance on C-Eval benchmark, where $*$ denotes chain-of-thought and Avg. represents average accuracy. We report the 5-shot and 0-shot performance with diagonal bar division.}
    \label{tab:c_eval_results}
\end{table}

\paragraph{Story Generation}
\label{subsubsec:story_gen}
\begin{wraptable}{r}{0.35\textwidth} 
    \small
    \centering
    \begin{tabular}{l c c}
        \toprule
        \bf Model & \bf \#Param. & \bf BBH \\
        \midrule
        LLaMA & 13B & \bf 37.1 \\
        \midrule
        ChatGLM & 6B & 31.3 \\
        Baichuan & 7B & 31.9 \\
        BatGPT & 15B & \underline{34.1} \\
        MOSS & 16B & 29.3 \\
        \midrule
        OpenBA & 15B & \underline{34.1} \\
        \bottomrule
    \end{tabular}
    \caption{Model performance on the BBH benchmark. We report the accuracy score for all the models.}
    \label{tab:bbh_results}
\end{wraptable} 
We evaluate the open-domain generation capability of OpenBA on the ROC dataset, where the model should continue generating based on the existing context and the story plot.
More concretely, we feed the model with the prompt directly and compare OpenBA with two other models: GPT-J~\citep{gpt-j} and OPT-13B~\citep{zhang2022opt}, which are also trained on the Pile corpus.
We randomly sample 100 generated cases and invite annotators to score the text from three aspects, including \textbf{coherence} between the generated text and the prompt, \textbf{consistency} and \textbf{correctness} of the generated text.
The annotators are allowed to choose "Tie" if it is hard to distinguish two generation cases.
As shown in Fig.~\ref{fig:humanevalres}, we can observe our model can obtain strong performance on the coherence and consistency aspect and attain comparable performance with two other models on the correctness aspect.

\subsection{Common Sense Reasoning}
\label{subsec:common_sense_reasoning}
We evaluate the common sense reasoning ability of OpenBA on four benchmarks, including MMLU, CMMLU, BBH, and C-Eval.
To ensure a fair comparison, we conduct all the evaluations under the held-out setting, follow the recommended setting of each benchmark, and compare with other strong LLMs under the same settings. 
For the MMLU~(Tab.~\ref{tab:mmlu_results}) and C-Eval~(Tab.~\ref{tab:c_eval_results}) benchmarks, we report the zero-shot, 5-shot, and 5-shot CoT results. 
For CMMLU~(Tab.~\ref{tab:cmmlu_results}), we report the  zero-shot, 5-shot, and zero-shot CoT results. 
We report the zero-shot results for BBH in Tab.~\ref{tab:bbh_results}.
It is worth noting that the first block for each table is multilingual- or English-oriented models, the second block is Chinese-oriented models, and we rank the models in each block by model size.
We can observe that, on all the benchmarks, OpenBA can achieve better performance than two strong Chinese-oriented models, i.e., ChatGLM~\citep{du2022glm} and BatGPT\footnote{BatGPT achieves the best performance on the official C-Eval leaderboard, but we do not obtain the reported results using the open-source version of BatGPT.}~\citep{li2023batgpt}, and obtain comparable results with Baichuan-7B model~\citep{Baichuan2023}, which is trained on datasets much larger than ours, i.e., 1.2TB tokens.
Furthermore, our model surpasses English-oriented models on most benchmarks and even outperforms some tasks where English-oriented models have over 100 billion parameters, e.g., BLOOM-176B, on the MMLU benchmark.
Additionally, OpenBA can achieve comparable scores under both zero-shot and few-shot settings and even performs slightly better under the zero-shot setting, indicating that the OpenBA model has a strong instruction-following capability.

\section{Analysis}
\label{sec:analysis}


\subsection{Model Architecture Selection}
\label{subsec:model_arch_selection}

Our asymmetric shallow-encoder deep-decoder model architecture stems from the following motivations and considerations:
\begin{itemize}
    \item \textbf{Enhanced Generative Capabilities.} For the three tasks in UL2, namely R-Denoising, S-Denoising, and X-Denoising, a deeper decoder setup is particularly effective for the S-Denoising task, which reflects the model's language modeling ability.
    \item \textbf{Potential Acceleration in Dialogue Inference.} Decoder-only architectures similar to GPT have already achieved excellent results in multi-turn dialogue tasks. However, for encoder-decoder models, how to store dialogue history presents a significant challenge. A common approach is to embed the dialogue history into the encoder's input. However, continuously altering this history results in increased computational costs in the encoder, and it's not amenable to acceleration via KV-caching. To address this challenge, we can place the dialogue history into the decoder. This shift imposes a greater demand on the decoder's capabilities. Thus, we explore training a deeper decoder to endow it with enhanced capabilities.

\end{itemize}

\begin{figure}[t]
\centering
\subfigure[]{\includegraphics[width=0.24\linewidth]{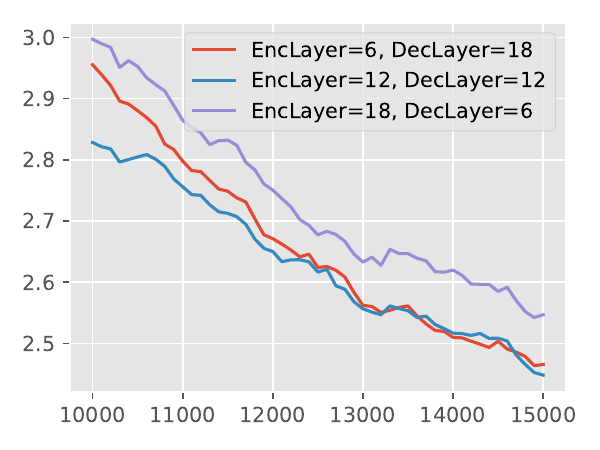}\label{fig:loss_R}}
\subfigure[]{\includegraphics[width=0.24\linewidth]{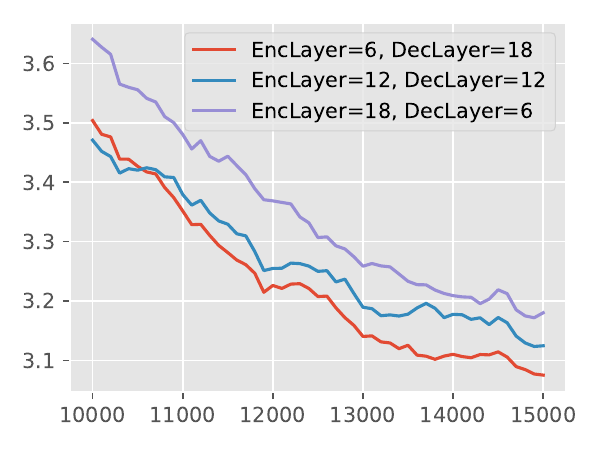}\label{fig:loss_S}}
\subfigure[]{\includegraphics[width=0.24\linewidth]{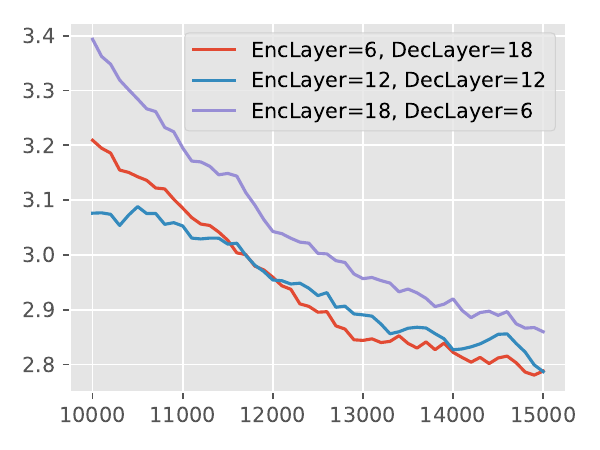}\label{fig:loss_X}}
\subfigure[]{\includegraphics[width=0.24\linewidth]{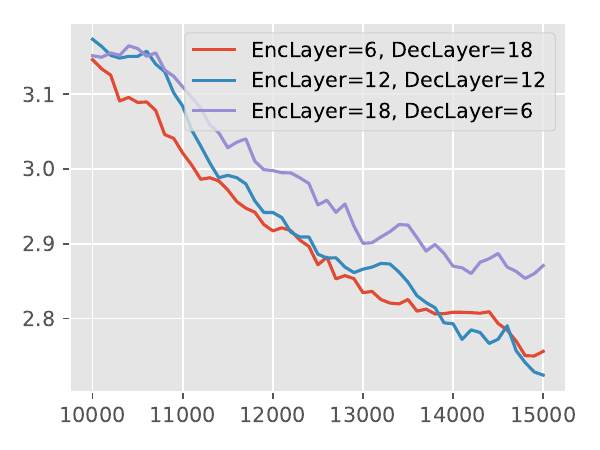}\label{fig:loss_mix}}
\\
\subfigure[]{\includegraphics[width=0.24\linewidth]{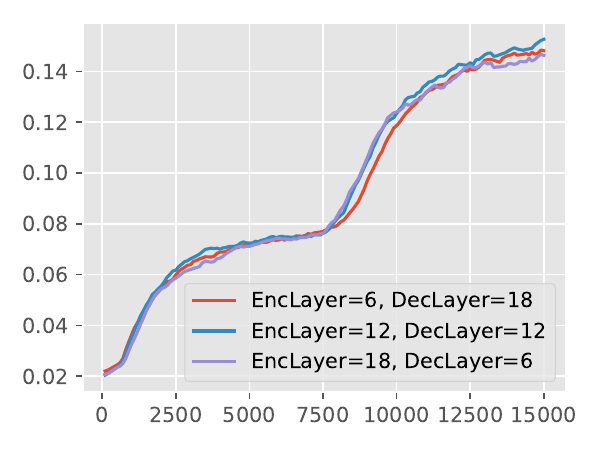}\label{fig:acc_R}}
\subfigure[]{\includegraphics[width=0.24\linewidth]{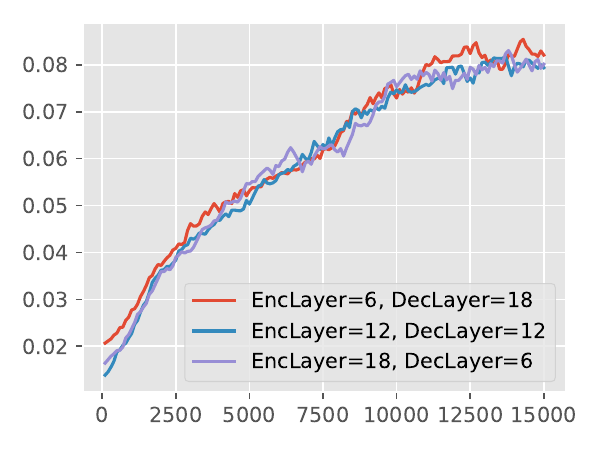}\label{fig:acc_S}}
\subfigure[]{\includegraphics[width=0.24\linewidth]{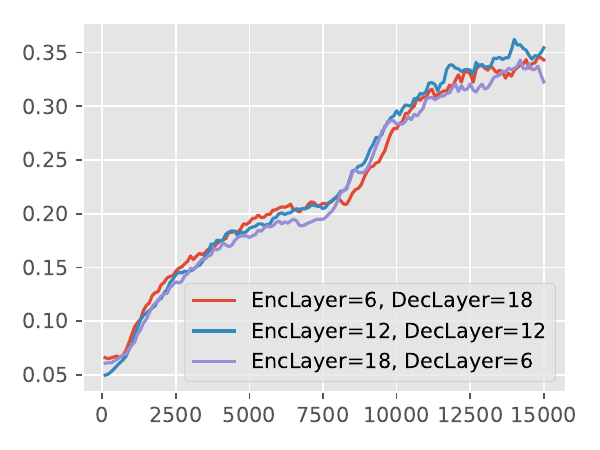}\label{fig:acc_X}}
\subfigure[]{\includegraphics[width=0.24\linewidth]{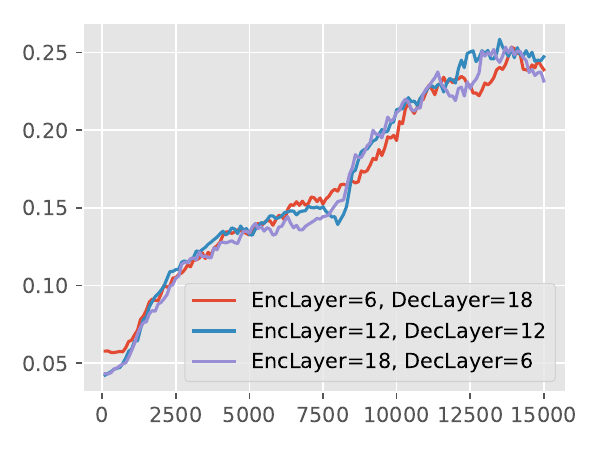}\label{fig:acc_mix}}
\caption{The performance in terms of loss and accuracy of the three model configurations across four denoising tasks. The first row of figures illustrates the loss performance, while the second row depicts the accuracy. The four columns respectively represent the four tasks: R-Denoising, S-Denoising, X-Denoising, and a combination of the three.}
\label{fig:lossacctask}
\end{figure}

We conduct experiments to explore the influence of the model architecture, where we train the model with the UL2 training objective.
Specifically, we set the batch size as 128 and the sequence length as 570/380.
We validate the model performance after 15k training steps.

\begin{figure}[t]
    \centering
\subfigure[MMLU]{
\includegraphics[width=0.48\linewidth]{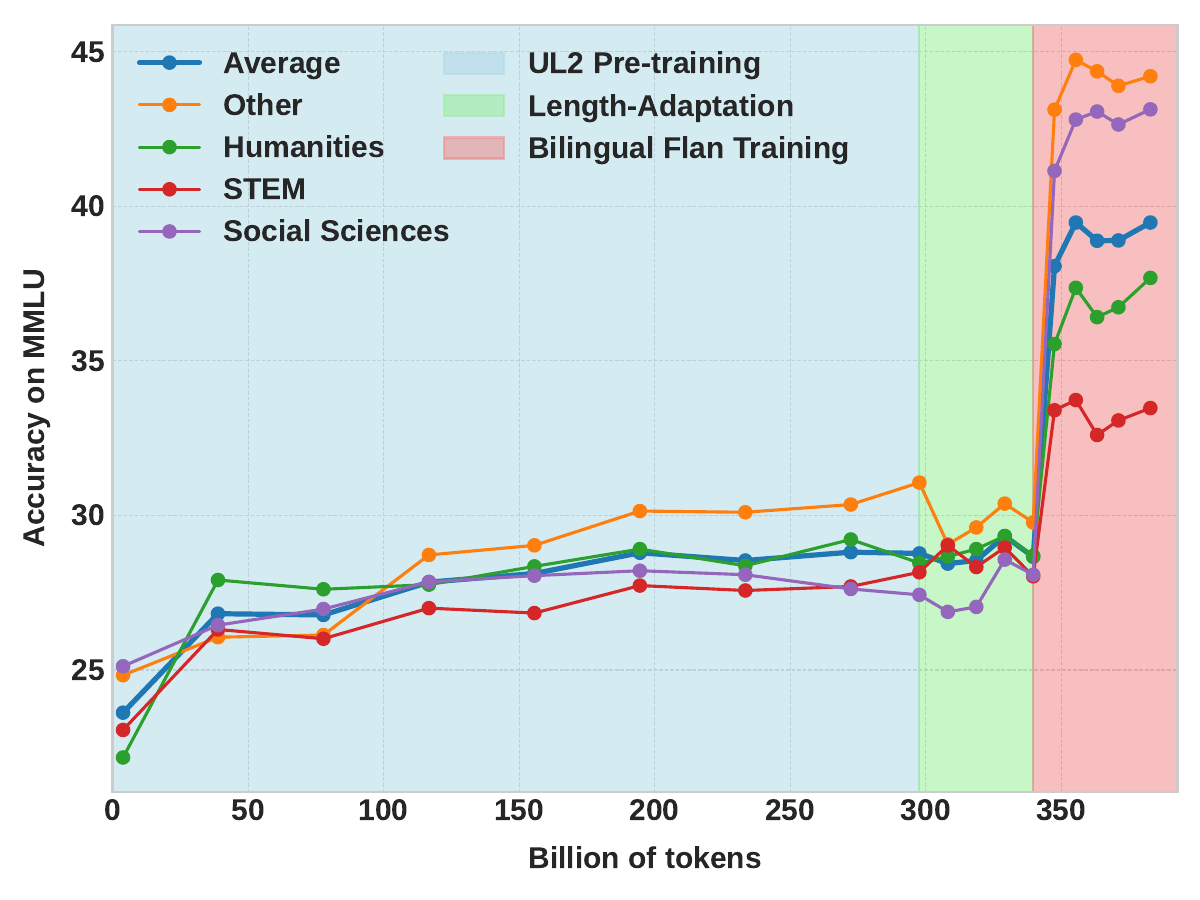}
\label{fig:fine_mmlus}
}
\hfill
\subfigure[CMMLU]{
\includegraphics[width=0.48\linewidth]{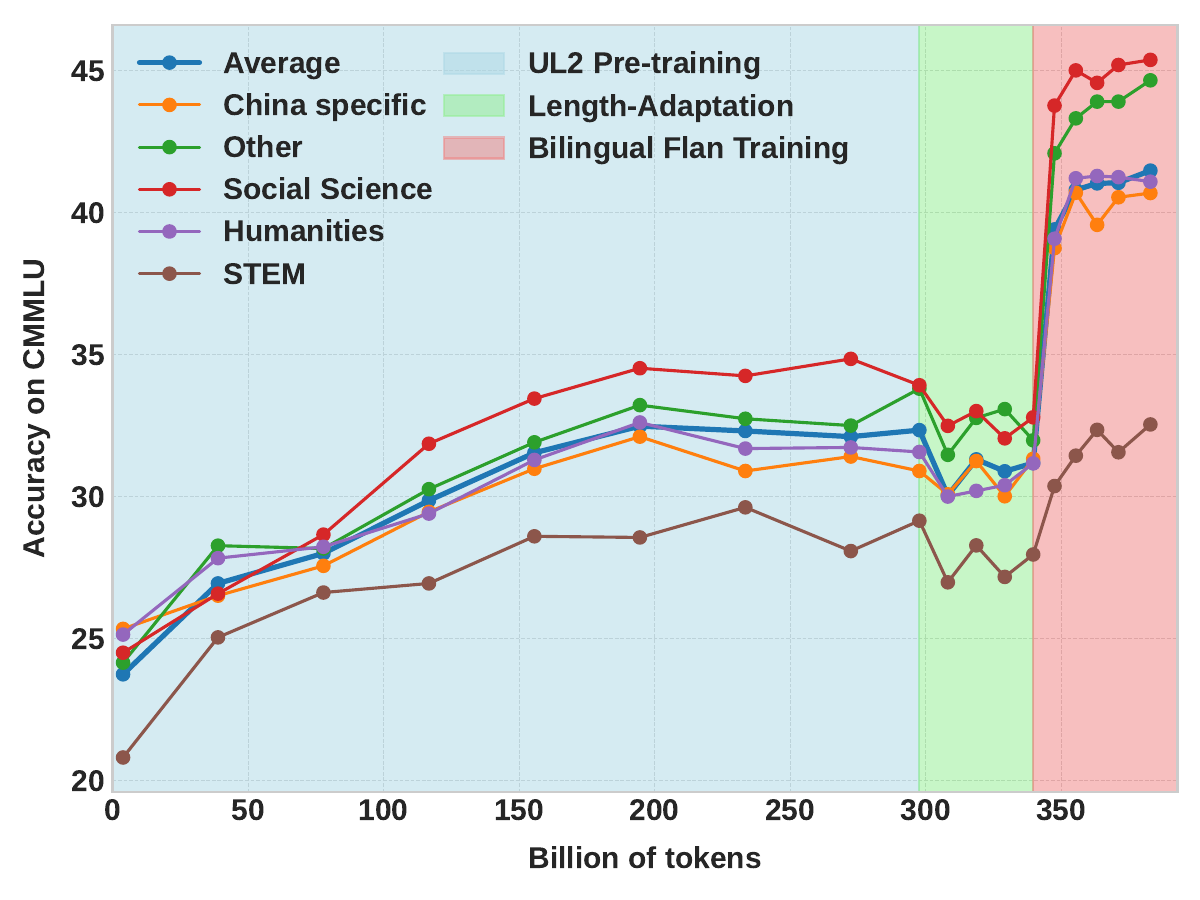}
\label{fig:fine_cmmlus}
}
\subfigure[BELEBELE]{
\includegraphics[width=0.48\linewidth]{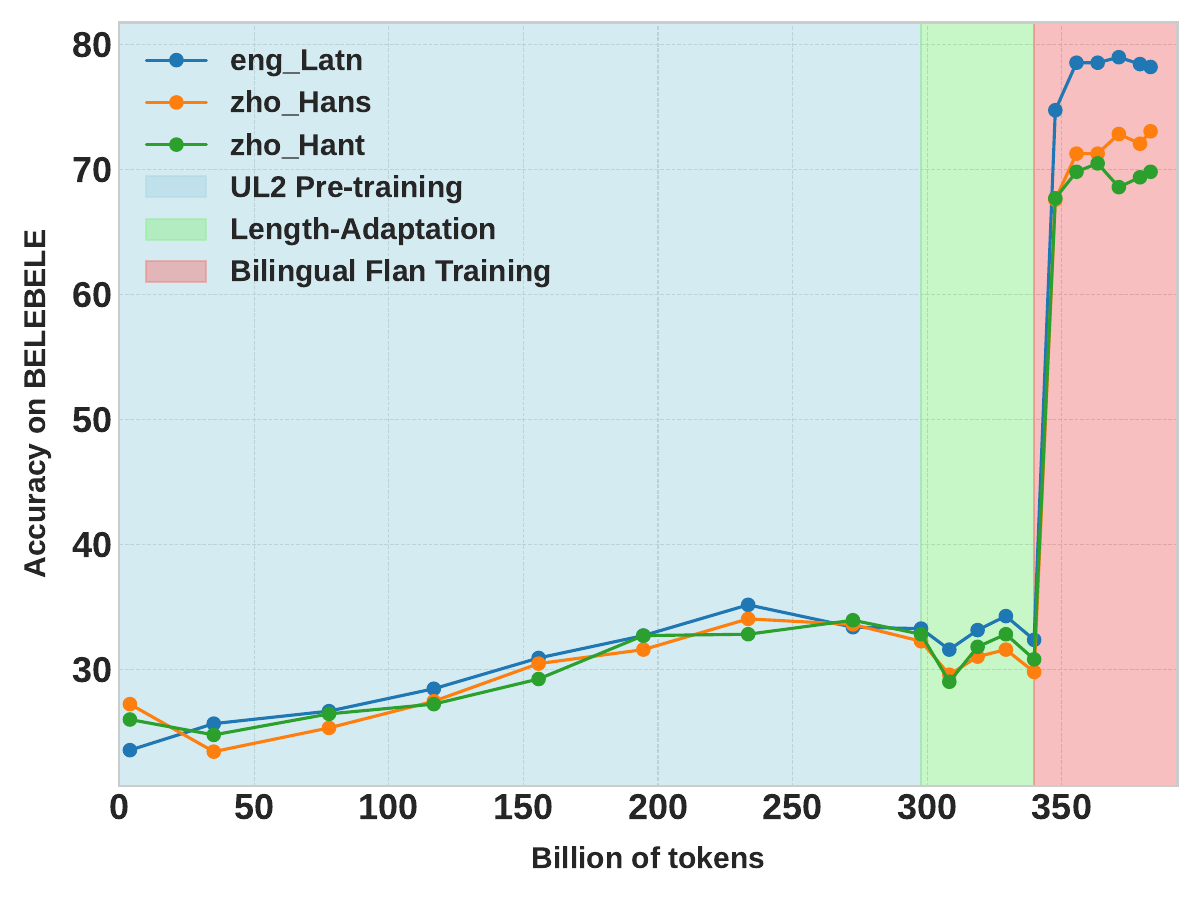}
\label{fig:fine_belebele}
}
\hfill
\subfigure[MMLU with different prefix tokens]{
\includegraphics[width=0.48\linewidth]{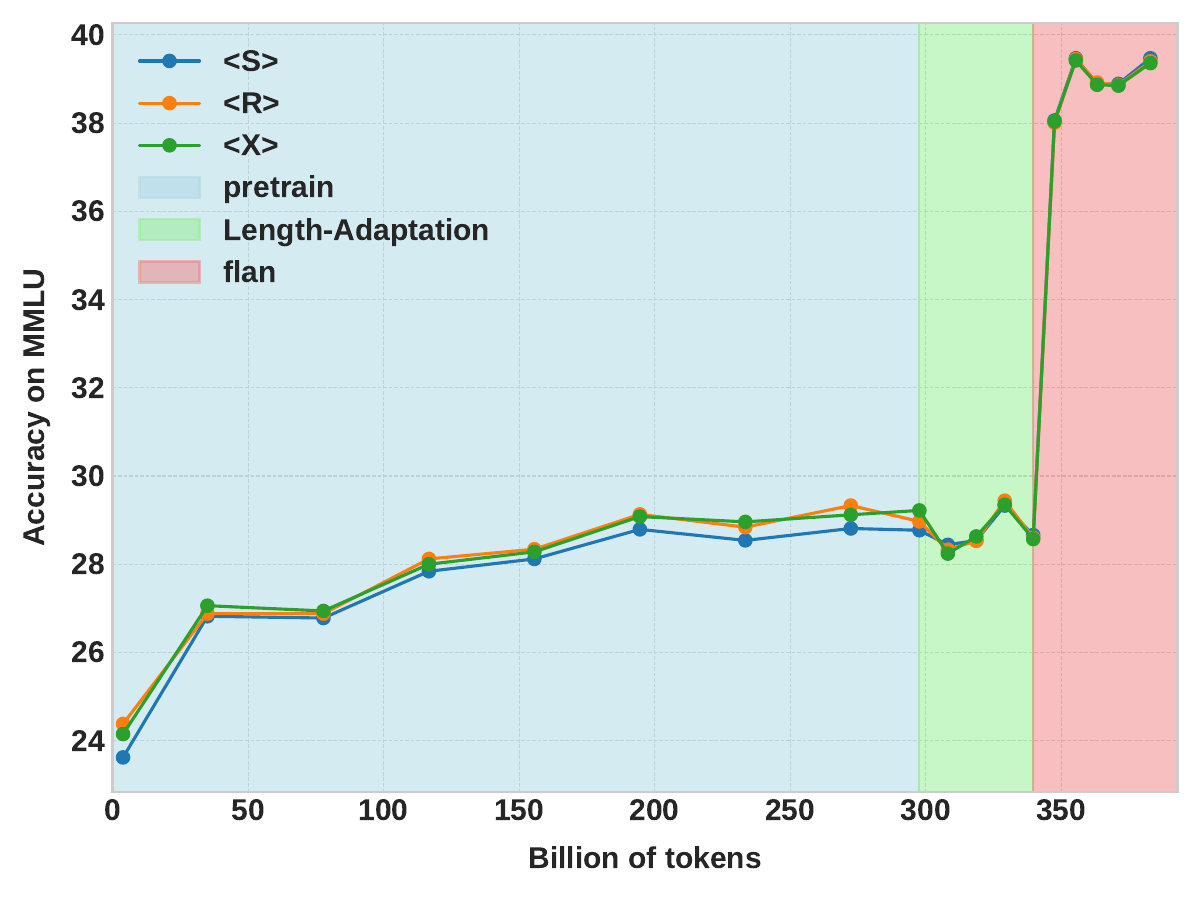}
\label{fig:mmlu_xrs}
}
    \caption{Evolution of model performance during training.}
    \label{fig:eval_during_training}
\end{figure}

\paragraph{Model Configuration}
We mainly explore three model structures: (1) a shallow encoder with a deep decoder, (2) a deep encoder with a shallow decoder, and (3) the encoder and decoder with equal depth.
We assess their performance metrics across the R-Denoising, S-Denoising, and X-Denoising tasks to learn their respective merits.
To maintain consistent parameter counts across different configurations, we adopt these layer structures: (1) EncoderLayer=18, DecoderLayer=6, (2) EncoderLayer=6, DecoderLayer=18, and (3) EncoderLayer=DecoderLayer=12.

\paragraph{Evaluation Metric}
To get a direct view of the model performance pre-trained from scratch, we choose Loss and Acc. as convenient metrics.
Specifically, we construct validation sets for R-Denoising, S-Denoising, X-Denoising, and a combination of the three, respectively, and test the model's performance throughout the training process. Acc. indicates the model's predictive accuracy for the next word:
\begin{equation}
    \text{Acc.} = \frac{1}{n}\sum\limits_{i = 1}^{n}\mathbb{I}(\text{argmax}_{w\in V}P(\boldsymbol{x}_i = w | \boldsymbol{x}_{<i}, \theta) = \boldsymbol{x}_i),
\end{equation}
where $n$ denotes the sequence length, $V$ denotes the vocabulary size and $\mathbb{I}$ is an indicator function.

\paragraph{Analysis}
Fig. \ref{fig:lossacctask} shows our results.
We can conclude that:
\begin{itemize}
    \item As a measurement of the model's generation ability, the S-Denoising task is generally more challenging to learn. This is evident as, regardless of the model configuration, the S-Denoising task consistently has a higher loss and a lower accuracy.
    \item The model with a shallow encoder and deep decoder configuration performs better on the S-denoising task (from Fig. \ref{fig:loss_S} and \ref{fig:acc_S}), though it doesn't outperform the balanced setup across all three tasks (from Fig. \ref{fig:loss_S} and \ref{fig:acc_S}).
\end{itemize}

\subsection{Evolution of Performance During Training}
\label{subsec:evolution_of_performance}
In this section, we evaluate the performance of OpenBA at various stages of the overall training.
We employ three benchmarks for evaluation, including MMLU for English common sense reasoning, CMMLU for Chinese common sense reasoning, and BELEBELE for reading comprehension. 
As shown in Fig. \ref{fig:fine_mmlus}, Fig. \ref{fig:fine_cmmlus}, and Fig. \ref{fig:fine_belebele}, the performance on most tasks increases with the number of training steps during the UL2 Pre-training stage, experiences slight fluctuations during the Length-Adaptation stage, and exhibits a significant improvement during the Bilingual Flan Training stage.
The emergence curves of Chinese and English are similar, indicating that our Bilingual Flan dataset effectively enhances multi-language task performance on held-out tasks.

Moreover, we measure the performance on MMLU when given different extra paradigm tokens, i.e., \{\texttt{<R>}, \texttt{<S>}, \texttt{<X>}\}.
We find that the performance with different extra paradigm tokens shows differences during the UL2 pre-training stage, while these differences gradually diminish in the subsequent stages.
This might be attributed to the fact that we utilize these extra paradigm tokens to guide the mode-switching only in the first stage for different UL2 tasks. 
Specifically, the performance for the S-denoising task in continuous writing is slightly inferior compared to the X-denoising and R-denoising tasks for masked span recovery.


\section{OpenBA-X: Downstream Task Adaptation}
\label{sec:downstream_task_adaptation}
After Stage III, we conduct supervised fine-tuning for OpenBA on four downstream tasks, including bilingual multi-turn dialogue (OpenBA-Chat), code generation (OpenBA-Code), instruction generation (OpenBA-InstructGen), and tool retrieval (OpenBA-Tool). In Section \ref{subsec:multi_turn_dialogue} to \ref{subsec:tool_retrieval}, we will provide details about the collection and processing of the downstream datasets. It is worth mentioning that we use the S-denoising strategy for fine-tuning all downstream tasks, i.e., adding the ``<S>'' token before each target text that is fed to the decoder.
We list all the instruction templates in Appendix \ref{sec:instruction_template}.

\begin{figure}[t]
\centering
\includegraphics[width=1.0\textwidth]{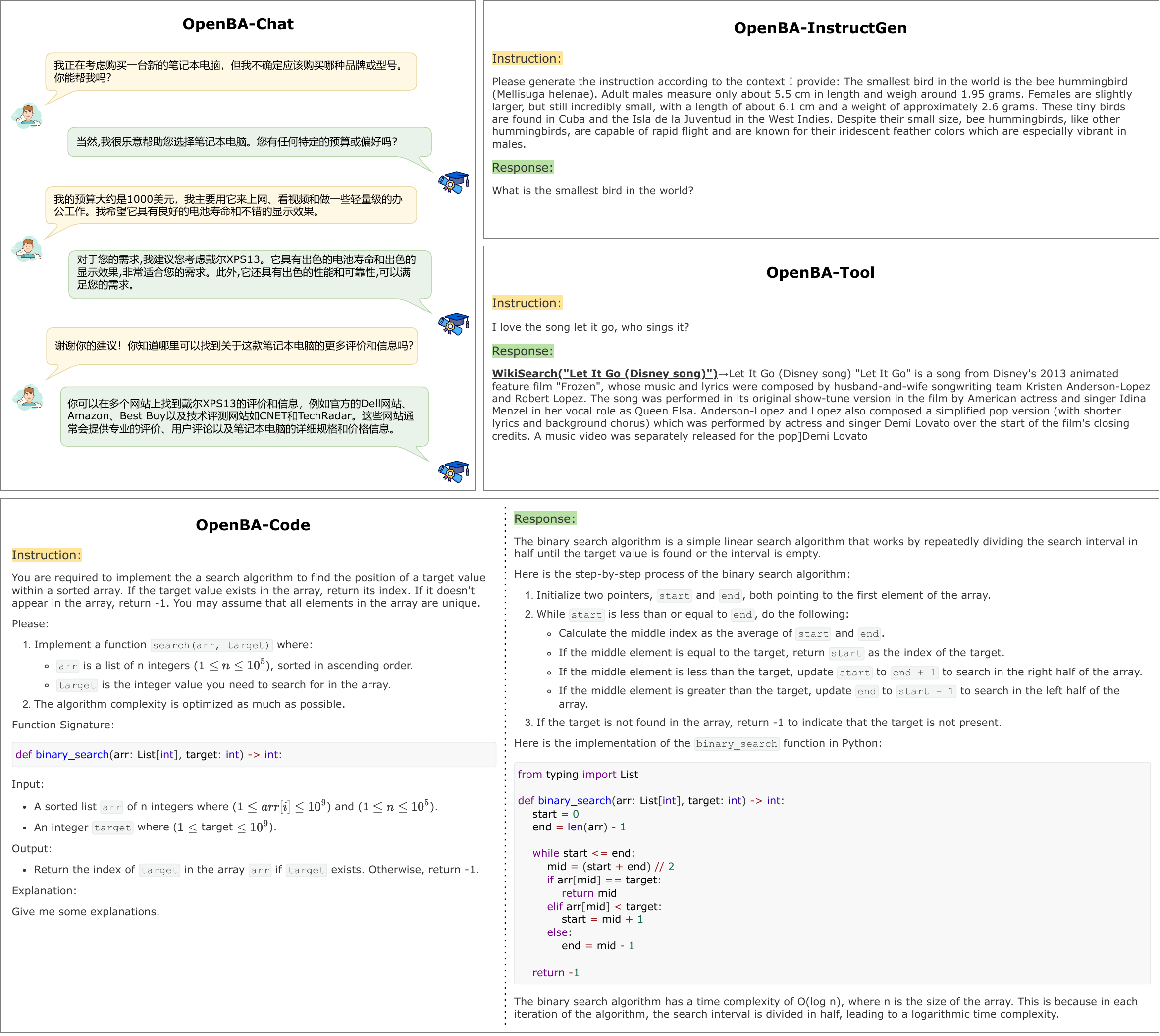}
\caption{Examples of OpenBA-X model on different downstream tasks. For the OpenBA-Chat model, we show the Chinese dialogue results. \textit{It is worth noting that there may be unrealistic content due to model hallucinations~\cite{rawte2023survey}.}}
\label{fig:downstream_task}
\end{figure}

\subsection{OpenBA-Chat: Bilingual Multi-turn Dialogue}
\label{subsec:multi_turn_dialogue}
\paragraph{Dataset Collection}
We build bilingual supervised multi-turn dialogue data from three distinct sources: DialogStudio~\citep{zhang2023dialogstudio}, BELLE~\citep{BELLE}, and ShareGPT\footnote{\url{https://huggingface.co/datasets/RyokoAI/ShareGPT52K}}. 
We use the DialogStudio dataset for English dialogue data as it contains diverse conversations for various scenarios.
As for Chinese dialogue data, we employ the BELLE dataset and ShareGPT data processed by others\footnote{\url{https://github.com/PhoebusSi/Alpaca-CoT}}.
We filter out the overly simple conversations based on their length, as well as the content containing model identity information, e.g., "I am ChatGPT."
More importantly, we manually annotate 40 bilingual conversations to identify OpenBA and repeat them ten times before adding them to the training dataset.

\paragraph{Dataset Processing}

Given $T$ turns of conversations involving two actors $H$ and $A$ in a dialogue, the data can be written as: $\boldsymbol{S} = (H_1, A_1, H_1, A_1, \cdots, H_t, A_t, \cdots, H_T, A_T),$
where ($H_{t}$, $A_{t}$) represents the $t$-th turn of the conversation. 
In order to enable the model to perceive the dialogue history and respond based on historical information, we process each dialogue data $\boldsymbol{S}$ into the set $\boldsymbol{D}$:
$$
\boldsymbol{D} = \bigcup_{t=1}^{T}\{\mathrm{Input}_t, \mathrm{Target}_t\}
               = \bigcup_{t=1}^{T}\{(H_1, A_1, H_2, A_2, \cdots, H_t), (A_t)\},
$$
where $\mathrm{Input}_t$ represents the input sequence, and $\mathrm{Output}_t$ represents the response sequence. 
The template to create the conversation for each instance is shown below:

\begin{sample}
\underline{Input:} ``Human: \{$H_0$\} Assistant: \{$A_0$\} $\cdots$ Human: \{$H_t$\} Assistant:'' \\
\underline{Output:} ``\{$A_t$\}''
\end{sample}

\subsection{OpenBA-Code: Code Generation}
\label{subsec:codegen}
\paragraph{Dataset Collection}
For code generation, we mainly focus on the Python language.
We choose a filtered version of the Evol-Instruct dataset~\citep{luo2023wizardcoder}, containing 26,588 code samples\footnote{\url{https://huggingface.co/datasets/mlabonne/Evol-Instruct-Python-26k}}.

\paragraph{Dataset Processing}
The original tokenizer of OpenBA would ignore consecutive spaces, thereby erasing the indentation information within the code. 
To tackle this issue, we incorporate three special tokens into the vocabulary: the tab character `\textbackslash t', the newline character `\textbackslash n', and consecutive spaces. 
We directly utilize the instructions from the original dataset as the instructions vary for different code contents. 

\subsection{OpenBA-InstructGen: Instruction Generation}
\label{subsec:instruct_gen}
\paragraph{Dataset Collection}
We construct a bilingual dataset for the instruction generation task by reversing the original instruction dataset~\citep{li2023self,alpaca}.
Specifically, we utilize the DollyV2 dataset~\citep{DatabricksBlog2023DollyV2}, 
Lima ~\citep{zhou2023lima} and its corresponding Chinese version Lima-Chinese\footnote{\url{https://huggingface.co/datasets/paralym/lima-chinese}}. 
More concretely, we repeat the Chinese corpus twice and combine them with the English dataset for language balance.

\paragraph{Dataset Processing}
Given an instruction ``$\mathrm{Instruction}$'' and its corresponding answer ``$\mathrm{Answer}$'', we utilize the following templates (including English and Chinese) to wrap each pair:

\begin{sample}
\underline{Input:} Please generate the instruction according to the text I provide:  
\{Answer\}. \\
\underline{Output:} \{Instruction\}. 
\end{sample}

\begin{sample}
\begin{CJK}{UTF8}{gbsn}
\underline{Input:} 请你根据提供的文本生成对应的指令：\{Answer\}。\\
\underline{Output:} \{Instruction\}。
\end{CJK}
\end{sample}

\subsection{OpenBA-Tool: Tool Retrieval}
\label{subsec:tool_retrieval}
\paragraph{Dataset Collection}
In order to enable the OpenBA model to respond to user instructions with the help of external tools~\citep{schick2023toolformer,wu2023visual}, we select Toolformer-Retrieval dataset\footnote{\url{https://huggingface.co/datasets/kentsui/open-toolformer-retrieval}}, which is designed for retrieval task.
For each instance, it is presented in the following format:
\begin{sample}
\underline{WikiSearch(\{Query Input\})} $\rightarrow$ \{Recalled Results\},
\end{sample}
where ``WikiSearch('' denotes the beginning of calling external tool (Wikipedia here), ``\{Query Input\}'' is the generated query input for the tool, and ``\{Recalled Results\}'' represents the results returned by invoking the tool.

\paragraph{Dataset Processing}
We utilize the instructions provided by the Toolformer-Retrieval dataset directly and discard the cases that fail to call tools. 
For simplicity, we use the model's output as a substitute for the actual retrieval result.

\section{Conclusion and Future Work}
\label{sec:con_future}
In this report, we present OpenBA, an Open-sourced 15B Bilingual Asymmetric seq2seq model pre-trained from scratch.
We provide all the necessary details to pre-train an asymmetric seq2seq model from scratch, including 1) how to construct and process the pre-training data, 2) how to construct the Bilingual Flan data collection, 3) the implementation details of model architectures, configurations, objectives, and training pipelines.
We also release our codes to supplement the descriptions of this report. 
On a variety of benchmarks, though fed with 380B tokens, OpenBA obtains remarkable performance, e.g., CMMLU and BELEBELE, and even surpasses the models consuming significantly more data.
\paragraph{Work in Progress}
We are currently working on the following directions about our model: \\
$\bullet$ We are conducting further evaluation to comprehensively calibrate the generation capability of OpenBA, especially for various tasks of controllable text generation~\citep{tang2023can}, and open-ended long text generation~\citep{liang2023open}. \\
$\bullet$ OpenBA faces ethical challenges and is prone to biases and toxicity since we have not yet performed any alignment operations~\citep{ouyang2022training}. After the alignment stage, we would like to test a few effective detoxification strategies on our model, e.g., detox-chain~\citep{tang2023detoxify}.\\
$\bullet$ \sout{The model's conversational capabilities need to be optimized for dialogue use cases}~\citep{yan2022deep}, \sout{such as the generation correctness}~\citep{tang2021chinese,bryant2022grammatical}. \\
$\bullet$ \sout{The ability to invoke tools, as we have tried to use sentinel tokens at the UL2 pre-training stage to activate different tool usage, i.e., multi-modal generation invoked by tools}~\cite{wu2023visual}.
\\
$\bullet$ OpenBA needs to be further extended in terms of input and output length to adapt to a wider range of tasks, such as dialogue generation.

\section*{Acknowledgments}
This work was supported by the National Key R\&D Program of China under Grant No. 2020AAA0108604, the National Science Foundation of China (NSFC No. 62206194 and No. 62106165), the Natural Science Foundation of Jiangsu Province, China (Grant No. BK20220488). We sincerely thank the GPU sponsor of the Supercomputing Center in Yancheng and technical advice from Bowen Yan and Jianye Hou.

\bibliographystyle{mybst}
\bibliography{en_cite}

\begin{thebibliography}{126}
\providecommand{\natexlab}[1]{#1}
\providecommand{\url}[1]{\texttt{#1}}
\expandafter\ifx\csname urlstyle\endcsname\relax
  \providecommand{\doi}[1]{doi: #1}\else
  \providecommand{\doi}{doi: \begingroup \urlstyle{rm}\Url}\fi

\bibitem[Artetxe et~al.(2019)Artetxe, Ruder, and Yogatama]{Artetxe:etal:2019}
Mikel Artetxe, Sebastian Ruder, and Dani Yogatama.
\newblock On the cross-lingual transferability of monolingual representations.
\newblock \emph{CoRR}, abs/1910.11856, 2019.

\bibitem[Bandarkar et~al.(2023)Bandarkar, Liang, Muller, Artetxe, Shukla, Husa,
  Goyal, Krishnan, Zettlemoyer, and Khabsa]{bandarkar2023belebele}
Lucas Bandarkar, Davis Liang, Benjamin Muller, Mikel Artetxe, Satya~Narayan
  Shukla, Donald Husa, Naman Goyal, Abhinandan Krishnan, Luke Zettlemoyer, and
  Madian Khabsa.
\newblock The belebele benchmark: a parallel reading comprehension dataset in
  122 language variants.
\newblock \emph{arXiv preprint arXiv:2308.16884}, 2023.

\bibitem[Barrault et~al.(2019)Barrault, Bojar, Costa-jussà, Federmann, Fishel,
  Graham, Haddow, Huck, Koehn, Malmasi, Monz, Müller, Pal, Post, and
  Zampieri]{WMT19}
Loïc Barrault, Ondřej Bojar, Marta~R. Costa-jussà, Christian Federmann, Mark
  Fishel, Yvette Graham, Barry Haddow, Matthias Huck, Philipp Koehn, Shervin
  Malmasi, Christof Monz, Mathias Müller, Santanu Pal, Matt Post, and Marcos
  Zampieri.
\newblock Findings of the 2019 conference on machine translation (wmt19).
\newblock In \emph{Proceedings of the Fourth Conference on Machine
  Translation}, volume~2, pp.\  1--61. Association for Computational
  Linguistics, 2019.

\bibitem[Barshan \& Fieguth(2015)Barshan and Fieguth]{barshan2015stage}
Elnaz Barshan and Paul Fieguth.
\newblock Stage-wise training: An improved feature learning strategy for deep
  models.
\newblock In \emph{Feature extraction: Modern questions and challenges}, pp.\
  49--59. PMLR, 2015.

\bibitem[Brown et~al.(2020)Brown, Mann, Ryder, Subbiah, Kaplan, Dhariwal,
  Neelakantan, Shyam, Sastry, Askell, et~al.]{brown2020language}
Tom Brown, Benjamin Mann, Nick Ryder, Melanie Subbiah, Jared~D Kaplan, Prafulla
  Dhariwal, Arvind Neelakantan, Pranav Shyam, Girish Sastry, Amanda Askell,
  et~al.
\newblock Language models are few-shot learners.
\newblock \emph{Advances in neural information processing systems},
  33:\penalty0 1877--1901, 2020.

\bibitem[Bryant et~al.(2022)Bryant, Yuan, Qorib, Cao, Ng, and
  Briscoe]{bryant2022grammatical}
Christopher Bryant, Zheng Yuan, Muhammad~Reza Qorib, Hannan Cao, Hwee~Tou Ng,
  and Ted Briscoe.
\newblock Grammatical error correction: A survey of the state of the art.
\newblock \emph{Computational Linguistics}, pp.\  1--59, 2022.

\bibitem[Bubeck et~al.(2023)Bubeck, Chandrasekaran, Eldan, Gehrke, Horvitz,
  Kamar, Lee, Lee, Li, Lundberg, et~al.]{bubeck2023sparks}
S{\'e}bastien Bubeck, Varun Chandrasekaran, Ronen Eldan, Johannes Gehrke, Eric
  Horvitz, Ece Kamar, Peter Lee, Yin~Tat Lee, Yuanzhi Li, Scott Lundberg,
  et~al.
\newblock Sparks of artificial general intelligence: Early experiments with
  gpt-4.
\newblock \emph{arXiv preprint arXiv:2303.12712}, 2023.

\bibitem[Chaudhary(2023)]{codealpaca}
Sahil Chaudhary.
\newblock Code alpaca: An instruction-following llama model for code
  generation.
\newblock \url{https://github.com/sahil280114/codealpaca}, 2023.

\bibitem[Chen et~al.(2023)Chen, Jiang, Chen, Wang, Yu, Chen, Zhang, Liang,
  Zhang, Zhang, et~al.]{chen2023phoenix}
Zhihong Chen, Feng Jiang, Junying Chen, Tiannan Wang, Fei Yu, Guiming Chen,
  Hongbo Zhang, Juhao Liang, Chen Zhang, Zhiyi Zhang, et~al.
\newblock Phoenix: Democratizing chatgpt across languages.
\newblock \emph{arXiv preprint arXiv:2304.10453}, 2023.

\bibitem[Chi et~al.(2020)Chi, Dong, Wei, Yang, Singhal, Wang, Song, Mao, Huang,
  and Zhou]{chi2020infoxlm}
Zewen Chi, Li~Dong, Furu Wei, Nan Yang, Saksham Singhal, Wenhui Wang, Xia Song,
  Xian-Ling Mao, Heyan Huang, and Ming Zhou.
\newblock Infoxlm: An information-theoretic framework for cross-lingual
  language model pre-training.
\newblock \emph{arXiv preprint arXiv:2007.07834}, 2020.

\bibitem[Chowdhery et~al.(2022)Chowdhery, Narang, Devlin, Bosma, Mishra,
  Roberts, Barham, Chung, Sutton, Gehrmann, et~al.]{chowdhery2022palm}
Aakanksha Chowdhery, Sharan Narang, Jacob Devlin, Maarten Bosma, Gaurav Mishra,
  Adam Roberts, Paul Barham, Hyung~Won Chung, Charles Sutton, Sebastian
  Gehrmann, et~al.
\newblock Palm: Scaling language modeling with pathways.
\newblock \emph{arXiv preprint arXiv:2204.02311}, 2022.

\bibitem[Chung et~al.(2022)Chung, Hou, Longpre, Zoph, Tay, Fedus, Li, Wang,
  Dehghani, Brahma, et~al.]{chung2022scaling}
Hyung~Won Chung, Le~Hou, Shayne Longpre, Barret Zoph, Yi~Tay, William Fedus,
  Eric Li, Xuezhi Wang, Mostafa Dehghani, Siddhartha Brahma, et~al.
\newblock Scaling instruction-finetuned language models.
\newblock \emph{arXiv preprint arXiv:2210.11416}, 2022.

\bibitem[Clark et~al.(2022)Clark, De~Las~Casas, Guy, Mensch, Paganini,
  Hoffmann, Damoc, Hechtman, Cai, Borgeaud, et~al.]{clark2022unified}
Aidan Clark, Diego De~Las~Casas, Aurelia Guy, Arthur Mensch, Michela Paganini,
  Jordan Hoffmann, Bogdan Damoc, Blake Hechtman, Trevor Cai, Sebastian
  Borgeaud, et~al.
\newblock Unified scaling laws for routed language models.
\newblock In \emph{International Conference on Machine Learning}, pp.\
  4057--4086. PMLR, 2022.

\bibitem[Conneau et~al.(2018)Conneau, Rinott, Lample, Williams, Bowman,
  Schwenk, and Stoyanov]{conneau2018xnli}
Alexis Conneau, Ruty Rinott, Guillaume Lample, Adina Williams, Samuel~R.
  Bowman, Holger Schwenk, and Veselin Stoyanov.
\newblock Xnli: Evaluating cross-lingual sentence representations.
\newblock In \emph{Proceedings of the 2018 Conference on Empirical Methods in
  Natural Language Processing}. Association for Computational Linguistics,
  2018.

\bibitem[Conover et~al.(2023{\natexlab{a}})Conover, Hayes, Mathur, Meng, Xie,
  Wan, Shah, Ghodsi, Wendell, Zaharia, et~al.]{conover2023free}
Mike Conover, Matt Hayes, Ankit Mathur, Xiangrui Meng, Jianwei Xie, Jun Wan,
  Sam Shah, Ali Ghodsi, Patrick Wendell, Matei Zaharia, et~al.
\newblock Free dolly: Introducing the world’s first truly open
  instruction-tuned llm, 2023{\natexlab{a}}.

\bibitem[Conover et~al.(2023{\natexlab{b}})Conover, Hayes, Mathur, Xie, Wan,
  Shah, Ghodsi, Wendell, Zaharia, and Xin]{DatabricksBlog2023DollyV2}
Mike Conover, Matt Hayes, Ankit Mathur, Jianwei Xie, Jun Wan, Sam Shah, Ali
  Ghodsi, Patrick Wendell, Matei Zaharia, and Reynold Xin.
\newblock Free dolly: Introducing the world's first truly open
  instruction-tuned llm, 2023{\natexlab{b}}.
\newblock URL
  \url{https://www.databricks.com/blog/2023/04/12/dolly-first-open-commercially-viable-instruction-tuned-llm}.

\bibitem[Coster \& Kauchak(2011)Coster and Kauchak]{coster2011simple}
William Coster and David Kauchak.
\newblock Simple english wikipedia: a new text simplification task.
\newblock In \emph{Proceedings of the 49th Annual Meeting of the Association
  for Computational Linguistics: Human Language Technologies}, pp.\  665--669,
  2011.

\bibitem[Cui et~al.(2019)Cui, Liu, Che, Xiao, Chen, Ma, Wang, and
  Hu]{cui-emnlp2019-cmrc2018}
Yiming Cui, Ting Liu, Wanxiang Che, Li~Xiao, Zhipeng Chen, Wentao Ma, Shijin
  Wang, and Guoping Hu.
\newblock A span-extraction dataset for {C}hinese machine reading
  comprehension.
\newblock In \emph{Proceedings of the 2019 Conference on Empirical Methods in
  Natural Language Processing and the 9th International Joint Conference on
  Natural Language Processing (EMNLP-IJCNLP)}, pp.\  5886--5891. Association
  for Computational Linguistics, 2019.

\bibitem[Cui et~al.(2023)Cui, Yang, and Yao]{cui2023efficient}
Yiming Cui, Ziqing Yang, and Xin Yao.
\newblock Efficient and effective text encoding for chinese llama and alpaca.
\newblock \emph{arXiv preprint arXiv:2304.08177}, 2023.

\bibitem[Devlin et~al.(2018)Devlin, Chang, Lee, and Toutanova]{devlin2018bert}
Jacob Devlin, Ming-Wei Chang, Kenton Lee, and Kristina Toutanova.
\newblock Bert: Pre-training of deep bidirectional transformers for language
  understanding.
\newblock \emph{arXiv preprint arXiv:1810.04805}, 2018.

\bibitem[Ding et~al.(2021)Ding, Yang, Hong, Zheng, Zhou, Yin, Lin, Zou, Shao,
  Yang, et~al.]{ding2021cogview}
Ming Ding, Zhuoyi Yang, Wenyi Hong, Wendi Zheng, Chang Zhou, Da~Yin, Junyang
  Lin, Xu~Zou, Zhou Shao, Hongxia Yang, et~al.
\newblock Cogview: Mastering text-to-image generation via transformers.
\newblock \emph{Advances in Neural Information Processing Systems},
  34:\penalty0 19822--19835, 2021.

\bibitem[Du et~al.(2022)Du, Qian, Liu, Ding, Qiu, Yang, and Tang]{du2022glm}
Zhengxiao Du, Yujie Qian, Xiao Liu, Ming Ding, Jiezhong Qiu, Zhilin Yang, and
  Jie Tang.
\newblock Glm: General language model pretraining with autoregressive blank
  infilling.
\newblock In \emph{Proceedings of the 60th Annual Meeting of the Association
  for Computational Linguistics (Volume 1: Long Papers)}, pp.\  320--335, 2022.

\bibitem[Duan(2016)]{duan2016nlpcc}
Nan Duan.
\newblock Overview of the nlpcc-iccpol 2016 shared task: Open domain chinese
  question answering.
\newblock In \emph{NLPCC/ICCPOL}, 2016.

\bibitem[Duh(2018)]{duh18multitarget}
Kevin Duh.
\newblock The multitarget ted talks task.
\newblock \url{http://www.cs.jhu.edu/~kevinduh/a/multitarget-tedtalks/}, 2018.

\bibitem[Faisal~Ladhak \& McKeown(2020)Faisal~Ladhak and
  McKeown]{ladhak-wiki-2020}
Claire~Cardie Faisal~Ladhak, Esin~Durmus and Kathleen McKeown.
\newblock Wikilingua: A new benchmark dataset for multilingual abstractive
  summarization.
\newblock In \emph{Findings of EMNLP, 2020}, 2020.

\bibitem[Fu et~al.(2023)Fu, Lam, Yu, So, Hu, Liu, and Collier]{fu2023decoder}
Zihao Fu, Wai Lam, Qian Yu, Anthony Man-Cho So, Shengding Hu, Zhiyuan Liu, and
  Nigel Collier.
\newblock Decoder-only or encoder-decoder? interpreting language model as a
  regularized encoder-decoder.
\newblock \emph{arXiv preprint arXiv:2304.04052}, 2023.

\bibitem[Gao et~al.(2020)Gao, Biderman, Black, Golding, Hoppe, Foster, Phang,
  He, Thite, Nabeshima, et~al.]{gao2020pile}
Leo Gao, Stella Biderman, Sid Black, Laurence Golding, Travis Hoppe, Charles
  Foster, Jason Phang, Horace He, Anish Thite, Noa Nabeshima, et~al.
\newblock The pile: An 800gb dataset of diverse text for language modeling.
\newblock \emph{arXiv preprint arXiv:2101.00027}, 2020.

\bibitem[Goyal et~al.(2022)Goyal, Gao, Chaudhary, Chen, Wenzek, Ju, Krishnan,
  Ranzato, Guzm{\'a}n, and Fan]{goyal2022flores}
Naman Goyal, Cynthia Gao, Vishrav Chaudhary, Peng-Jen Chen, Guillaume Wenzek,
  Da~Ju, Sanjana Krishnan, Marc’Aurelio Ranzato, Francisco Guzm{\'a}n, and
  Angela Fan.
\newblock The flores-101 evaluation benchmark for low-resource and multilingual
  machine translation.
\newblock \emph{Transactions of the Association for Computational Linguistics},
  10:\penalty0 522--538, 2022.

\bibitem[Grun(2010)]{grun2010introduction}
Paul Grun.
\newblock Introduction to infiniband for end users.
\newblock \emph{White paper, InfiniBand Trade Association}, 55, 2010.

\bibitem[Guan et~al.(2021)Guan, Mao, Fan, Liu, Ding, and Huang]{guan2021long}
Jian Guan, Xiaoxi Mao, Changjie Fan, Zitao Liu, Wenbiao Ding, and Minlie Huang.
\newblock Long text generation by modeling sentence-level and discourse-level
  coherence.
\newblock \emph{arXiv preprint arXiv:2105.08963}, 2021.

\bibitem[Hendrycks et~al.(2020)Hendrycks, Burns, Basart, Zou, Mazeika, Song,
  and Steinhardt]{hendrycks2020measuring}
Dan Hendrycks, Collin Burns, Steven Basart, Andy Zou, Mantas Mazeika, Dawn
  Song, and Jacob Steinhardt.
\newblock Measuring massive multitask language understanding.
\newblock \emph{arXiv preprint arXiv:2009.03300}, 2020.

\bibitem[Hoffmann et~al.(2022)Hoffmann, Borgeaud, Mensch, Buchatskaya, Cai,
  Rutherford, Casas, Hendricks, Welbl, Clark, et~al.]{hoffmann2022training}
Jordan Hoffmann, Sebastian Borgeaud, Arthur Mensch, Elena Buchatskaya, Trevor
  Cai, Eliza Rutherford, Diego de~Las Casas, Lisa~Anne Hendricks, Johannes
  Welbl, Aidan Clark, et~al.
\newblock Training compute-optimal large language models.
\newblock \emph{arXiv preprint arXiv:2203.15556}, 2022.

\bibitem[Hu et~al.(2020)Hu, Richardson, Xu, Li, Kübler, and Moss]{OCNLI}
Hai Hu, Kyle Richardson, Liang Xu, Lu~Li, Sandra Kübler, and Lawrence Moss.
\newblock Ocnli: Original chinese natural language inference.
\newblock In \emph{Findings of the Association for Computational Linguistics},
  pp.\  3512–3526. Association for Computational Linguistics, 2020.

\bibitem[Hu et~al.(2022)Hu, Guo, Wu, Liu, Wen, and Yu]{CHEF}
Xuming Hu, Zhijiang Guo, GuanYu Wu, Aiwei Liu, Lijie Wen, and Philip Yu.
\newblock Chef: A pilot chinese dataset for evidence-based fact-checking.
\newblock In \emph{Proceedings of the 2022 Conference of the North American
  Chapter of the Association for Computational Linguistics: Human Language
  Technologies}, pp.\  3362–3376. Association for Computational Linguistics,
  2022.

\bibitem[Huang et~al.(2023)Huang, Bai, Zhu, Zhang, Zhang, Su, Liu, Lv, Zhang,
  Lei, et~al.]{huang2023c}
Yuzhen Huang, Yuzhuo Bai, Zhihao Zhu, Junlei Zhang, Jinghan Zhang, Tangjun Su,
  Junteng Liu, Chuancheng Lv, Yikai Zhang, Jiayi Lei, et~al.
\newblock C-eval: A multi-level multi-discipline chinese evaluation suite for
  foundation models.
\newblock \emph{arXiv preprint arXiv:2305.08322}, 2023.

\bibitem[Inc.(2023)]{Baichuan2023}
Baichuan Inc.
\newblock Baichuan-7b.
\newblock \url{https://github.com/baichuan-inc/Baichuan-7B}, 2023.

\bibitem[Ji et~al.(2023)Ji, Deng, Gong, Peng, Niu, Ma, and Li]{BELLE}
Yunjie Ji, Yong Deng, Yan Gong, Yiping Peng, Qiang Niu, Baochang Ma, and
  Xiangang Li.
\newblock Belle: Be everyone's large language model engine.
\newblock \url{https://github.com/LianjiaTech/BELLE}, 2023.

\bibitem[Jiahua~Liu \& Sun(2019)Jiahua~Liu and Sun]{XQA}
Zhiyuan~Liu Jiahua~Liu, Yankai~Lin and Maosong Sun.
\newblock Xqa: A cross-lingual open-domain question answering dataset.
\newblock In \emph{Proceedings of the 57th Annual Meeting of the Association
  for Computational Linguistics}, pp.\  2358–2368, 2019.

\bibitem[Kaplan et~al.(2020)Kaplan, McCandlish, Henighan, Brown, Chess, Child,
  Gray, Radford, Wu, and Amodei]{kaplan2020scaling}
Jared Kaplan, Sam McCandlish, Tom Henighan, Tom~B Brown, Benjamin Chess, Rewon
  Child, Scott Gray, Alec Radford, Jeffrey Wu, and Dario Amodei.
\newblock Scaling laws for neural language models.
\newblock \emph{arXiv preprint arXiv:2001.08361}, 2020.

\bibitem[Lewis et~al.(2019)Lewis, Liu, Goyal, Ghazvininejad, Mohamed, Levy,
  Stoyanov, and Zettlemoyer]{lewis2019bart}
Mike Lewis, Yinhan Liu, Naman Goyal, Marjan Ghazvininejad, Abdelrahman Mohamed,
  Omer Levy, Ves Stoyanov, and Luke Zettlemoyer.
\newblock Bart: Denoising sequence-to-sequence pre-training for natural
  language generation, translation, and comprehension.
\newblock \emph{arXiv preprint arXiv:1910.13461}, 2019.

\bibitem[Li et~al.(2023{\natexlab{a}})Li, Zhang, Koto, Yang, Zhao, Gong, Duan,
  and Baldwin]{li2023cmmlu}
Haonan Li, Yixuan Zhang, Fajri Koto, Yifei Yang, Hai Zhao, Yeyun Gong, Nan
  Duan, and Timothy Baldwin.
\newblock Cmmlu: Measuring massive multitask language understanding in chinese.
\newblock \emph{arXiv preprint arXiv:2306.09212}, 2023{\natexlab{a}}.

\bibitem[Li et~al.(2015)Li, Galley, Brockett, Gao, and Dolan]{li2015diversity}
Jiwei Li, Michel Galley, Chris Brockett, Jianfeng Gao, and Bill Dolan.
\newblock A diversity-promoting objective function for neural conversation
  models.
\newblock \emph{arXiv preprint arXiv:1510.03055}, 2015.

\bibitem[Li et~al.(2019)Li, He, Shi, Jiang, Liang, jiang, Zhang, Lyu, and
  Zhu]{DuIE2.0}
Shuangjie Li, Wei He, Yabing Shi, Wenbin Jiang, Haijin Liang, Ye~jiang, Yang
  Zhang, Yajuan Lyu, and Yong Zhu.
\newblock Duie: A large-scale chinese dataset for information extraction.
\newblock In \emph{Tang, J., Kan, MY., Zhao, D., Li, S., Zan, H. (eds) Natural
  Language Processing and Chinese Computing}, volume 11839. Springer, Cham,
  2019.

\bibitem[Li et~al.(2023{\natexlab{b}})Li, Yu, Zhou, Schick, Zettlemoyer, Levy,
  Weston, and Lewis]{li2023self}
Xian Li, Ping Yu, Chunting Zhou, Timo Schick, Luke Zettlemoyer, Omer Levy,
  Jason Weston, and Mike Lewis.
\newblock Self-alignment with instruction backtranslation.
\newblock \emph{arXiv preprint arXiv:2308.06259}, 2023{\natexlab{b}}.

\bibitem[Li et~al.(2022)Li, Zhang, Zhao, Shen, Liu, Mao, and Zhang]{CSL}
Yudong Li, Yuqing Zhang, Zhe Zhao, Linlin Shen, Weijie Liu, Weiquan Mao, and
  Hui Zhang.
\newblock Csl: A large-scale chinese scientific literature dataset.
\newblock In \emph{Proceedings of the 29th International Conference on
  Computational Linguistics}, pp.\  3917–3923. International Committee on
  Computational Linguistics, 2022.

\bibitem[Li et~al.(2023{\natexlab{c}})Li, Zhang, Zhao, Yang, and
  Yang]{li2023batgpt}
Zuchao Li, Shitou Zhang, Hai Zhao, Yifei Yang, and Dongjie Yang.
\newblock Batgpt: A bidirectional autoregessive talker from generative
  pre-trained transformer.
\newblock \emph{arXiv preprint arXiv:2307.00360}, 2023{\natexlab{c}}.

\bibitem[Liang et~al.(2023{\natexlab{a}})Liang, Gonen, Mao, Hou, Goyal,
  Ghazvininejad, Zettlemoyer, and Khabsa]{liang2023xlm}
Davis Liang, Hila Gonen, Yuning Mao, Rui Hou, Naman Goyal, Marjan
  Ghazvininejad, Luke Zettlemoyer, and Madian Khabsa.
\newblock Xlm-v: Overcoming the vocabulary bottleneck in multilingual masked
  language models.
\newblock \emph{arXiv preprint arXiv:2301.10472}, 2023{\natexlab{a}}.

\bibitem[Liang et~al.(2023{\natexlab{b}})Liang, Tang, Li, and
  Zhang]{liang2023open}
Xiaobo Liang, Zecheng Tang, Juntao Li, and Min Zhang.
\newblock Open-ended long text generation via masked language modeling.
\newblock In \emph{Proceedings of the 61st Annual Meeting of the Association
  for Computational Linguistics (Volume 1: Long Papers)}, pp.\  223--241,
  2023{\natexlab{b}}.

\bibitem[Lin(2004)]{lin2004rouge}
Chin-Yew Lin.
\newblock Rouge: A package for automatic evaluation of summaries.
\newblock In \emph{Text summarization branches out}, pp.\  74--81, 2004.

\bibitem[Liu et~al.(2020)Liu, Zhang, Chen, Cao, and Li]{liu2020clts}
Xiaojun Liu, Chuang Zhang, Xiaojun Chen, Yanan Cao, and Jinpeng Li.
\newblock Clts: a new chinese long text summarization dataset.
\newblock In \emph{CCF International Conference on Natural Language Processing
  and Chinese Computing}, pp.\  531--542. Springer, 2020.

\bibitem[Longpre et~al.(2023)Longpre, Hou, Vu, Webson, Chung, Tay, Zhou, Le,
  Zoph, Wei, et~al.]{longpre2023flan}
Shayne Longpre, Le~Hou, Tu~Vu, Albert Webson, Hyung~Won Chung, Yi~Tay, Denny
  Zhou, Quoc~V Le, Barret Zoph, Jason Wei, et~al.
\newblock The flan collection: Designing data and methods for effective
  instruction tuning.
\newblock \emph{arXiv preprint arXiv:2301.13688}, 2023.

\bibitem[Luo et~al.(2023)Luo, Xu, Zhao, Sun, Geng, Hu, Tao, Ma, Lin, and
  Jiang]{luo2023wizardcoder}
Ziyang Luo, Can Xu, Pu~Zhao, Qingfeng Sun, Xiubo Geng, Wenxiang Hu, Chongyang
  Tao, Jing Ma, Qingwei Lin, and Daxin Jiang.
\newblock Wizardcoder: Empowering code large language models with
  evol-instruct.
\newblock \emph{arXiv preprint arXiv:2306.08568}, 2023.

\bibitem[Lv et~al.(2022)Lv, Cao, Geng, Ai, Yan, and Fu]{MD-SCS}
Qi~Lv, Ziqiang Cao, Lei Geng, Chunhui Ai, Xu~Yan, and Guohong Fu.
\newblock General and domain adaptive chinese spelling check with error
  consistent pretraining.
\newblock In \emph{ACM Trans. Asian Low-Resour. Lang. Inf. Process}.
  Association for Computing Machinery, 2022.

\bibitem[Min et~al.(2021)Min, Lewis, Zettlemoyer, and
  Hajishirzi]{min2021metaicl}
Sewon Min, Mike Lewis, Luke Zettlemoyer, and Hannaneh Hajishirzi.
\newblock Metaicl: Learning to learn in context.
\newblock \emph{arXiv preprint arXiv:2110.15943}, 2021.

\bibitem[Mostafazadeh et~al.(2016)Mostafazadeh, Chambers, He, Parikh, Batra,
  Vanderwende, Kohli, and Allen]{mostafazadeh2016corpus}
Nasrin Mostafazadeh, Nathanael Chambers, Xiaodong He, Devi Parikh, Dhruv Batra,
  Lucy Vanderwende, Pushmeet Kohli, and James Allen.
\newblock A corpus and cloze evaluation for deeper understanding of commonsense
  stories.
\newblock In \emph{Proceedings of the 2016 Conference of the North American
  Chapter of the Association for Computational Linguistics: Human Language
  Technologies}, pp.\  839--849, 2016.

\bibitem[Nijkamp et~al.(2022)Nijkamp, Pang, Hayashi, Tu, Wang, Zhou, Savarese,
  and Xiong]{nijkamp2022codegen}
Erik Nijkamp, Bo~Pang, Hiroaki Hayashi, Lifu Tu, Huan Wang, Yingbo Zhou, Silvio
  Savarese, and Caiming Xiong.
\newblock Codegen: An open large language model for code with multi-turn
  program synthesis.
\newblock \emph{arXiv preprint arXiv:2203.13474}, 2022.

\bibitem[Nye et~al.(2021)Nye, Andreassen, Gur-Ari, Michalewski, Austin, Bieber,
  Dohan, Lewkowycz, Bosma, Luan, et~al.]{nye2021show}
Maxwell Nye, Anders~Johan Andreassen, Guy Gur-Ari, Henryk Michalewski, Jacob
  Austin, David Bieber, David Dohan, Aitor Lewkowycz, Maarten Bosma, David
  Luan, et~al.
\newblock Show your work: Scratchpads for intermediate computation with
  language models.
\newblock \emph{arXiv preprint arXiv:2112.00114}, 2021.

\bibitem[Ouyang et~al.(2022)Ouyang, Wu, Jiang, Almeida, Wainwright, Mishkin,
  Zhang, Agarwal, Slama, Ray, et~al.]{ouyang2022training}
Long Ouyang, Jeffrey Wu, Xu~Jiang, Diogo Almeida, Carroll Wainwright, Pamela
  Mishkin, Chong Zhang, Sandhini Agarwal, Katarina Slama, Alex Ray, et~al.
\newblock Training language models to follow instructions with human feedback.
\newblock \emph{Advances in Neural Information Processing Systems},
  35:\penalty0 27730--27744, 2022.

\bibitem[Pan et~al.(2020)Pan, Zhang, Ji, and Yang]{pan2020privacy}
Xudong Pan, Mi~Zhang, Shouling Ji, and Min Yang.
\newblock Privacy risks of general-purpose language models.
\newblock In \emph{2020 IEEE Symposium on Security and Privacy (SP)}, pp.\
  1314--1331. IEEE, 2020.

\bibitem[Penedo et~al.(2023{\natexlab{a}})Penedo, Malartic, Hesslow, Cojocaru,
  Cappelli, Alobeidli, Pannier, Almazrouei, and Launay]{penedo2023refinedweb}
Guilherme Penedo, Quentin Malartic, Daniel Hesslow, Ruxandra Cojocaru,
  Alessandro Cappelli, Hamza Alobeidli, Baptiste Pannier, Ebtesam Almazrouei,
  and Julien Launay.
\newblock The refinedweb dataset for falcon llm: outperforming curated corpora
  with web data, and web data only.
\newblock \emph{arXiv preprint arXiv:2306.01116}, 2023{\natexlab{a}}.

\bibitem[Penedo et~al.(2023{\natexlab{b}})Penedo, Malartic, Hesslow, Cojocaru,
  Cappelli, Alobeidli, Pannier, Almazrouei, and Launay]{refinedweb}
Guilherme Penedo, Quentin Malartic, Daniel Hesslow, Ruxandra Cojocaru,
  Alessandro Cappelli, Hamza Alobeidli, Baptiste Pannier, Ebtesam Almazrouei,
  and Julien Launay.
\newblock The {R}efined{W}eb dataset for {F}alcon {LLM}: outperforming curated
  corpora with web data, and web data only.
\newblock \emph{arXiv preprint arXiv:2306.01116}, 2023{\natexlab{b}}.
\newblock URL \url{https://arxiv.org/abs/2306.01116}.

\bibitem[Pillutla et~al.(2021)Pillutla, Swayamdipta, Zellers, Thickstun,
  Welleck, Choi, and Harchaoui]{pillutla2021mauve}
Krishna Pillutla, Swabha Swayamdipta, Rowan Zellers, John Thickstun, Sean
  Welleck, Yejin Choi, and Zaid Harchaoui.
\newblock Mauve: Measuring the gap between neural text and human text using
  divergence frontiers.
\newblock \emph{Advances in Neural Information Processing Systems},
  34:\penalty0 4816--4828, 2021.

\bibitem[Post(2018)]{post-2018-call}
Matt Post.
\newblock A call for clarity in reporting {BLEU} scores.
\newblock In \emph{Proceedings of the Third Conference on Machine Translation:
  Research Papers}, pp.\  186--191, Belgium, Brussels, October 2018.
  Association for Computational Linguistics.
\newblock URL \url{https://www.aclweb.org/anthology/W18-6319}.

\bibitem[Radford et~al.(2018)Radford, Narasimhan, Salimans, Sutskever,
  et~al.]{radford2018improving}
Alec Radford, Karthik Narasimhan, Tim Salimans, Ilya Sutskever, et~al.
\newblock Improving language understanding by generative pre-training.
\newblock 2018.

\bibitem[Radford et~al.(2019)Radford, Wu, Child, Luan, Amodei, Sutskever,
  et~al.]{radford2019language}
Alec Radford, Jeffrey Wu, Rewon Child, David Luan, Dario Amodei, Ilya
  Sutskever, et~al.
\newblock Language models are unsupervised multitask learners.
\newblock \emph{OpenAI blog}, 1\penalty0 (8):\penalty0 9, 2019.

\bibitem[Rae et~al.(2021)Rae, Borgeaud, Cai, Millican, Hoffmann, Song,
  Aslanides, Henderson, Ring, Young, et~al.]{rae2021scaling}
Jack~W Rae, Sebastian Borgeaud, Trevor Cai, Katie Millican, Jordan Hoffmann,
  Francis Song, John Aslanides, Sarah Henderson, Roman Ring, Susannah Young,
  et~al.
\newblock Scaling language models: Methods, analysis \& insights from training
  gopher.
\newblock \emph{arXiv preprint arXiv:2112.11446}, 2021.

\bibitem[Raffel et~al.(2020)Raffel, Shazeer, Roberts, Lee, Narang, Matena,
  Zhou, Li, and Liu]{raffel2020exploring}
Colin Raffel, Noam Shazeer, Adam Roberts, Katherine Lee, Sharan Narang, Michael
  Matena, Yanqi Zhou, Wei Li, and Peter~J Liu.
\newblock Exploring the limits of transfer learning with a unified text-to-text
  transformer.
\newblock \emph{The Journal of Machine Learning Research}, 21\penalty0
  (1):\penalty0 5485--5551, 2020.

\bibitem[Rajbhandari et~al.(2020)Rajbhandari, Rasley, Ruwase, and
  He]{rajbhandari2020zero}
Samyam Rajbhandari, Jeff Rasley, Olatunji Ruwase, and Yuxiong He.
\newblock Zero: Memory optimizations toward training trillion parameter models.
\newblock In \emph{SC20: International Conference for High Performance
  Computing, Networking, Storage and Analysis}, pp.\  1--16. IEEE, 2020.

\bibitem[Rawte et~al.(2023)Rawte, Sheth, and Das]{rawte2023survey}
Vipula Rawte, Amit Sheth, and Amitava Das.
\newblock A survey of hallucination in large foundation models.
\newblock \emph{arXiv preprint arXiv:2309.05922}, 2023.

\bibitem[Sanh et~al.(2021)Sanh, Webson, Raffel, Bach, Sutawika, Alyafeai,
  Chaffin, Stiegler, Scao, Raja, et~al.]{sanh2021multitask}
Victor Sanh, Albert Webson, Colin Raffel, Stephen~H Bach, Lintang Sutawika,
  Zaid Alyafeai, Antoine Chaffin, Arnaud Stiegler, Teven~Le Scao, Arun Raja,
  et~al.
\newblock Multitask prompted training enables zero-shot task generalization.
\newblock \emph{arXiv preprint arXiv:2110.08207}, 2021.

\bibitem[Scao et~al.(2022)Scao, Fan, Akiki, Pavlick, Ili{\'c}, Hesslow,
  Castagn{\'e}, Luccioni, Yvon, Gall{\'e}, et~al.]{scao2022bloom}
Teven~Le Scao, Angela Fan, Christopher Akiki, Ellie Pavlick, Suzana Ili{\'c},
  Daniel Hesslow, Roman Castagn{\'e}, Alexandra~Sasha Luccioni, Fran{\c{c}}ois
  Yvon, Matthias Gall{\'e}, et~al.
\newblock Bloom: A 176b-parameter open-access multilingual language model.
\newblock \emph{arXiv preprint arXiv:2211.05100}, 2022.

\bibitem[Schaller(1997)]{schaller1997moore}
Robert~R Schaller.
\newblock Moore's law: past, present and future.
\newblock \emph{IEEE spectrum}, 34\penalty0 (6):\penalty0 52--59, 1997.

\bibitem[Schick et~al.(2023)Schick, Dwivedi-Yu, Dess{\`\i}, Raileanu, Lomeli,
  Zettlemoyer, Cancedda, and Scialom]{schick2023toolformer}
Timo Schick, Jane Dwivedi-Yu, Roberto Dess{\`\i}, Roberta Raileanu, Maria
  Lomeli, Luke Zettlemoyer, Nicola Cancedda, and Thomas Scialom.
\newblock Toolformer: Language models can teach themselves to use tools.
\newblock \emph{arXiv preprint arXiv:2302.04761}, 2023.

\bibitem[Shao et~al.(2018)Shao, Liu, Lai, Tseng, and Tsai]{DRCD}
Chih~Chieh Shao, Trois Liu, Yuting Lai, Yiying Tseng, and Sam Tsai.
\newblock Drcd: a chinese machine reading comprehension dataset.
\newblock \emph{arXiv preprint arXiv:1806.00920}, 2018.

\bibitem[Shao et~al.(2019{\natexlab{a}})Shao, Huang, Wen, Xu, and
  Zhu]{AdvertiseGen}
Zhihong Shao, Minlie Huang, Jiangtao Wen, Wenfei Xu, and Xiaoyan Zhu.
\newblock Long and diverse text generation with planning-based hierarchical
  variational model.
\newblock In \emph{Proceedings of the 2019 Conference on Empirical Methods in
  Natural Language Processing and the 9th International Joint Conference on
  Natural Language Processing}, pp.\  3257–3268. Association for
  Computational Linguistics, 2019{\natexlab{a}}.

\bibitem[Shao et~al.(2019{\natexlab{b}})Shao, Huang, Wen, Xu, and
  Zhu]{shao2019long}
Zhihong Shao, Minlie Huang, Jiangtao Wen, Wenfei Xu, and Xiaoyan Zhu.
\newblock Long and diverse text generation with planning-based hierarchical
  variational model.
\newblock \emph{arXiv preprint arXiv:1908.06605}, 2019{\natexlab{b}}.

\bibitem[Shazeer(2020)]{shazeer2020glu}
Noam Shazeer.
\newblock Glu variants improve transformer.
\newblock \emph{arXiv preprint arXiv:2002.05202}, 2020.

\bibitem[Shen et~al.(2022)Shen, Wang, Chen, Xiao, Liu, and Wu]{DuExplain}
Yaozong Shen, Lijie Wang, Ying Chen, Xinyan Xiao, Jing Liu, and Hua Wu.
\newblock An interpretability evaluation benchmark for pre-trained language
  model.
\newblock \emph{arXiv preprint arXiv:2207.13948}, 2022.

\bibitem[Shoeybi et~al.(2019)Shoeybi, Patwary, Puri, LeGresley, Casper, and
  Catanzaro]{shoeybi2019megatron}
Mohammad Shoeybi, Mostofa Patwary, Raul Puri, Patrick LeGresley, Jared Casper,
  and Bryan Catanzaro.
\newblock Megatron-lm: Training multi-billion parameter language models using
  model parallelism.
\newblock \emph{arXiv preprint arXiv:1909.08053}, 2019.

\bibitem[Smith et~al.(2022)Smith, Patwary, Norick, LeGresley, Rajbhandari,
  Casper, Liu, Prabhumoye, Zerveas, Korthikanti, et~al.]{smith2022using}
Shaden Smith, Mostofa Patwary, Brandon Norick, Patrick LeGresley, Samyam
  Rajbhandari, Jared Casper, Zhun Liu, Shrimai Prabhumoye, George Zerveas,
  Vijay Korthikanti, et~al.
\newblock Using deepspeed and megatron to train megatron-turing nlg 530b, a
  large-scale generative language model.
\newblock \emph{arXiv preprint arXiv:2201.11990}, 2022.

\bibitem[Soltan et~al.(2022)Soltan, Ananthakrishnan, FitzGerald, Gupta, Hamza,
  Khan, Peris, Rawls, Rosenbaum, Rumshisky, et~al.]{soltan2022alexatm}
Saleh Soltan, Shankar Ananthakrishnan, Jack FitzGerald, Rahul Gupta, Wael
  Hamza, Haidar Khan, Charith Peris, Stephen Rawls, Andy Rosenbaum, Anna
  Rumshisky, et~al.
\newblock Alexatm 20b: Few-shot learning using a large-scale multilingual
  seq2seq model.
\newblock \emph{arXiv preprint arXiv:2208.01448}, 2022.

\bibitem[Su et~al.(2021)Su, Lu, Pan, Murtadha, Wen, and Liu]{su2021roformer}
Jianlin Su, Yu~Lu, Shengfeng Pan, Ahmed Murtadha, Bo~Wen, and Yunfeng Liu.
\newblock Roformer: Enhanced transformer with rotary position embedding.
\newblock \emph{arXiv preprint arXiv:2104.09864}, 2021.

\bibitem[Sun et~al.(2023)Sun, Zhang, He, Li, Cheng, Yan, Liu, Shao, Tang, Zhao,
  Chen, Zheng, Zhou, Li, Zhan, Zhou, Li, Yang, Wu, Yin, Huang, and
  Qiu]{sun2023moss}
Tianxiang Sun, Xiaotian Zhang, Zhengfu He, Peng Li, Qinyuan Cheng, Hang Yan,
  Xiangyang Liu, Yunfan Shao, Qiong Tang, Xingjian Zhao, Ke~Chen, Yining Zheng,
  Zhejian Zhou, Ruixiao Li, Jun Zhan, Yunhua Zhou, Linyang Li, Xiaogui Yang,
  Lingling Wu, Zhangyue Yin, Xuanjing Huang, and Xipeng Qiu.
\newblock Moss: Training conversational language models from synthetic data.
\newblock 2023.

\bibitem[Sun et~al.(2021)Sun, Wang, Feng, Ding, Pang, Shang, Liu, Chen, Zhao,
  Lu, et~al.]{sun2021ernie}
Yu~Sun, Shuohuan Wang, Shikun Feng, Siyu Ding, Chao Pang, Junyuan Shang,
  Jiaxiang Liu, Xuyi Chen, Yanbin Zhao, Yuxiang Lu, et~al.
\newblock Ernie 3.0: Large-scale knowledge enhanced pre-training for language
  understanding and generation.
\newblock \emph{arXiv preprint arXiv:2107.02137}, 2021.

\bibitem[Suzgun et~al.(2022)Suzgun, Scales, Sch{\"a}rli, Gehrmann, Tay, Chung,
  Chowdhery, Le, Chi, Zhou, et~al.]{suzgun2022challenging}
Mirac Suzgun, Nathan Scales, Nathanael Sch{\"a}rli, Sebastian Gehrmann, Yi~Tay,
  Hyung~Won Chung, Aakanksha Chowdhery, Quoc~V Le, Ed~H Chi, Denny Zhou, et~al.
\newblock Challenging big-bench tasks and whether chain-of-thought can solve
  them.
\newblock \emph{arXiv preprint arXiv:2210.09261}, 2022.

\bibitem[Tang et~al.(2021{\natexlab{a}})Tang, Li, Liu, Hong, Wu, and
  Wang]{DuReader}
Hongxuan Tang, Hongyu Li, Jing Liu, Yu~Hong, Hua Wu, and Haifeng Wang.
\newblock Dureader\_robust: A chinese dataset towards evaluating robustness and
  generalization of machine reading comprehension in real-world applications.
\newblock In \emph{Proceedings of the 59th Annual Meeting of the Association
  for Computational Linguistics and the 11th International Joint Conference on
  Natural Language Processing}, volume~2, pp.\  955–963, 2021{\natexlab{a}}.

\bibitem[Tang et~al.(2021{\natexlab{b}})Tang, Ji, Zhao, and
  Li]{tang2021chinese}
Zecheng Tang, Yixin Ji, Yibo Zhao, and Junhui Li.
\newblock Chinese grammatical error correction enhanced by data augmentation
  from word and character levels.
\newblock In \emph{Proceedings of the 20th Chinese National Conference on
  Computational Linguistics, Hohhot, China}, pp.\  13--15, 2021{\natexlab{b}}.

\bibitem[Tang et~al.(2023{\natexlab{a}})Tang, Wang, Zhou, Li, Cao, and
  Zhang]{tang2023can}
Zecheng Tang, Pinzheng Wang, Keyan Zhou, Juntao Li, Ziqiang Cao, and Min Zhang.
\newblock Can diffusion model achieve better performance in text generation?
  bridging the gap between training and inference!
\newblock \emph{arXiv preprint arXiv:2305.04465}, 2023{\natexlab{a}}.

\bibitem[Tang et~al.(2023{\natexlab{b}})Tang, Zhou, Wang, Ding, Li,
  et~al.]{tang2023detoxify}
Zecheng Tang, Keyan Zhou, Pinzheng Wang, Yuyang Ding, Juntao Li, et~al.
\newblock Detoxify language model step-by-step.
\newblock \emph{arXiv preprint arXiv:2308.08295}, 2023{\natexlab{b}}.

\bibitem[Taori et~al.(2023)Taori, Gulrajani, Zhang, Dubois, Li, Guestrin,
  Liang, and Hashimoto]{alpaca}
Rohan Taori, Ishaan Gulrajani, Tianyi Zhang, Yann Dubois, Xuechen Li, Carlos
  Guestrin, Percy Liang, and Tatsunori~B. Hashimoto.
\newblock Stanford alpaca: An instruction-following llama model.
\newblock \url{https://github.com/tatsu-lab/stanford_alpaca}, 2023.

\bibitem[Tay et~al.(2022)Tay, Dehghani, Tran, Garcia, Wei, Wang, Chung, Bahri,
  Schuster, Zheng, et~al.]{tay2022ul2}
Yi~Tay, Mostafa Dehghani, Vinh~Q Tran, Xavier Garcia, Jason Wei, Xuezhi Wang,
  Hyung~Won Chung, Dara Bahri, Tal Schuster, Steven Zheng, et~al.
\newblock Ul2: Unifying language learning paradigms.
\newblock In \emph{The Eleventh International Conference on Learning
  Representations}, 2022.

\bibitem[Touvron et~al.(2023{\natexlab{a}})Touvron, Lavril, Izacard, Martinet,
  Lachaux, Lacroix, Rozi{\`e}re, Goyal, Hambro, Azhar,
  et~al.]{touvron2023llama}
Hugo Touvron, Thibaut Lavril, Gautier Izacard, Xavier Martinet, Marie-Anne
  Lachaux, Timoth{\'e}e Lacroix, Baptiste Rozi{\`e}re, Naman Goyal, Eric
  Hambro, Faisal Azhar, et~al.
\newblock Llama: Open and efficient foundation language models.
\newblock \emph{arXiv preprint arXiv:2302.13971}, 2023{\natexlab{a}}.

\bibitem[Touvron et~al.(2023{\natexlab{b}})Touvron, Martin, Stone, Albert,
  Almahairi, Babaei, Bashlykov, Batra, Bhargava, Bhosale,
  et~al.]{touvron2023llama2}
Hugo Touvron, Louis Martin, Kevin Stone, Peter Albert, Amjad Almahairi, Yasmine
  Babaei, Nikolay Bashlykov, Soumya Batra, Prajjwal Bhargava, Shruti Bhosale,
  et~al.
\newblock Llama 2: Open foundation and fine-tuned chat models.
\newblock \emph{arXiv preprint arXiv:2307.09288}, 2023{\natexlab{b}}.

\bibitem[Vaswani et~al.(2017)Vaswani, Shazeer, Parmar, Uszkoreit, Jones, Gomez,
  Kaiser, and Polosukhin]{vaswani2017attention}
Ashish Vaswani, Noam Shazeer, Niki Parmar, Jakob Uszkoreit, Llion Jones,
  Aidan~N Gomez, {\L}ukasz Kaiser, and Illia Polosukhin.
\newblock Attention is all you need.
\newblock \emph{Advances in neural information processing systems}, 30, 2017.

\bibitem[Wang et~al.(2019)Wang, Pruksachatkun, Nangia, Singh, Michael, Hill,
  Levy, and Bowman]{wang2019superglue}
Alex Wang, Yada Pruksachatkun, Nikita Nangia, Amanpreet Singh, Julian Michael,
  Felix Hill, Omer Levy, and Samuel Bowman.
\newblock Superglue: A stickier benchmark for general-purpose language
  understanding systems.
\newblock \emph{Advances in neural information processing systems}, 32, 2019.

\bibitem[Wang \& Komatsuzaki(2021)Wang and Komatsuzaki]{gpt-j}
Ben Wang and Aran Komatsuzaki.
\newblock {GPT-J-6B: A 6 Billion Parameter Autoregressive Language Model}.
\newblock \url{https://github.com/kingoflolz/mesh-transformer-jax}, May 2021.

\bibitem[Wang et~al.(2023)Wang, Liu, Xi, Qiang, Zhao, Qin, and
  Liu]{wang2023huatuo}
Haochun Wang, Chi Liu, Nuwa Xi, Zewen Qiang, Sendong Zhao, Bing Qin, and Ting
  Liu.
\newblock Huatuo: Tuning llama model with chinese medical knowledge.
\newblock \emph{arXiv preprint arXiv:2304.06975}, 2023.

\bibitem[Wang et~al.(2022)Wang, Wei, Schuurmans, Le, Chi, Narang, Chowdhery,
  and Zhou]{wang2022self}
Xuezhi Wang, Jason Wei, Dale Schuurmans, Quoc Le, Ed~Chi, Sharan Narang,
  Aakanksha Chowdhery, and Denny Zhou.
\newblock Self-consistency improves chain of thought reasoning in language
  models.
\newblock \emph{arXiv preprint arXiv:2203.11171}, 2022.

\bibitem[Wang et~al.(2017)Wang, Liu, and Shi]{Math23K}
Yan Wang, Xiaojiang Liu, and Shuming Shi.
\newblock Deep neural solver for math word problems.
\newblock In \emph{Proceedings of the 2017 Conference on Empirical Methods in
  Natural Language Processing}, pp.\  845–854. Association for Computational
  Linguistics, 2017.

\bibitem[Wei et~al.(2021)Wei, Bosma, Zhao, Guu, Yu, Lester, Du, Dai, and
  Le]{wei2021finetuned}
Jason Wei, Maarten Bosma, Vincent~Y Zhao, Kelvin Guu, Adams~Wei Yu, Brian
  Lester, Nan Du, Andrew~M Dai, and Quoc~V Le.
\newblock Finetuned language models are zero-shot learners.
\newblock \emph{arXiv preprint arXiv:2109.01652}, 2021.

\bibitem[Wei et~al.(2022)Wei, Wang, Schuurmans, Bosma, Chi, Le, and
  Zhou]{wei2022chain}
Jason Wei, Xuezhi Wang, Dale Schuurmans, Maarten Bosma, Ed~Chi, Quoc Le, and
  Denny Zhou.
\newblock Chain of thought prompting elicits reasoning in large language
  models.
\newblock \emph{arXiv preprint arXiv:2201.11903}, 2022.

\bibitem[Wu et~al.(2022)Wu, Raghavendra, Gupta, Acun, Ardalani, Maeng, Chang,
  Aga, Huang, Bai, et~al.]{wu2022sustainable}
Carole-Jean Wu, Ramya Raghavendra, Udit Gupta, Bilge Acun, Newsha Ardalani,
  Kiwan Maeng, Gloria Chang, Fiona Aga, Jinshi Huang, Charles Bai, et~al.
\newblock Sustainable ai: Environmental implications, challenges and
  opportunities.
\newblock \emph{Proceedings of Machine Learning and Systems}, 4:\penalty0
  795--813, 2022.

\bibitem[Wu et~al.(2023{\natexlab{a}})Wu, Yin, Qi, Wang, Tang, and
  Duan]{wu2023visual}
Chenfei Wu, Shengming Yin, Weizhen Qi, Xiaodong Wang, Zecheng Tang, and Nan
  Duan.
\newblock Visual chatgpt: Talking, drawing and editing with visual foundation
  models.
\newblock \emph{arXiv preprint arXiv:2303.04671}, 2023{\natexlab{a}}.

\bibitem[Wu et~al.(2023{\natexlab{b}})Wu, Zhao, Li, Qin, and
  Xiong]{wu2023improving}
Yang Wu, Yanyan Zhao, Zhongyang Li, Bing Qin, and Kai Xiong.
\newblock Improving cross-task generalization with step-by-step instructions.
\newblock \emph{arXiv preprint arXiv:2305.04429}, 2023{\natexlab{b}}.

\bibitem[Wu et~al.(2017)Wu, Wu, Xing, Zhou, and Li]{douban}
Yu~Wu, Wei Wu, Chen Xing, Ming Zhou, and Zhoujun Li.
\newblock Sequential matching network: A new archtechture for multi-turn
  response selection in retrieval-based chatbots.
\newblock In \emph{Proceedings of the 55th Annual Meeting of the Association
  for Computational Linguistics}, volume~1, pp.\  496–505. Association for
  Computational Linguistics, 2017.

\bibitem[Xu(2019)]{bright_xu_2019_3402023}
Bright Xu.
\newblock Nlp chinese corpus: Large scale chinese corpus for nlp, September
  2019.
\newblock URL \url{https://doi.org/10.5281/zenodo.3402023}.

\bibitem[Xu et~al.(2023)Xu, Sun, Zheng, Geng, Zhao, Feng, Tao, and
  Jiang]{xu2023wizardlm}
Can Xu, Qingfeng Sun, Kai Zheng, Xiubo Geng, Pu~Zhao, Jiazhan Feng, Chongyang
  Tao, and Daxin Jiang.
\newblock Wizardlm: Empowering large language models to follow complex
  instructions.
\newblock \emph{arXiv preprint arXiv:2304.12244}, 2023.

\bibitem[Xu et~al.(2020)Xu, Hu, Zhang, Li, Cao, Li, Xu, Sun, Yu, Yu, Tian,
  Dong, Liu, Shi, Cui, Li, Zeng, Wang, Xie, Li, Patterson, Tian, Zhang, Zhou,
  Liu, Zhao, Zhao, Yue, Zhang, Yang, Richardson, and Lan]{xu-etal-2020-clue}
Liang Xu, Hai Hu, Xuanwei Zhang, Lu~Li, Chenjie Cao, Yudong Li, Yechen Xu, Kai
  Sun, Dian Yu, Cong Yu, Yin Tian, Qianqian Dong, Weitang Liu, Bo~Shi, Yiming
  Cui, Junyi Li, Jun Zeng, Rongzhao Wang, Weijian Xie, Yanting Li, Yina
  Patterson, Zuoyu Tian, Yiwen Zhang, He~Zhou, Shaoweihua Liu, Zhe Zhao, Qipeng
  Zhao, Cong Yue, Xinrui Zhang, Zhengliang Yang, Kyle Richardson, and Zhenzhong
  Lan.
\newblock Clue: A {C}hinese language understanding evaluation benchmark.
\newblock In \emph{Proceedings of the 28th International Conference on
  Computational Linguistics}, pp.\  4762--4772. International Committee on
  Computational Linguistics, 2020.

\bibitem[Xu et~al.(2021)Xu, Lu, Yuan, Zhang, Xu, Yuan, Wei, Pan, Tian, Qin, and
  Hai]{FewCLUE}
Liang Xu, Xiaojing Lu, Chenyang Yuan, Xuanwei Zhang, Huilin Xu, Hu~Yuan, Guoao
  Wei, Xiang Pan, Xin Tian, Libo Qin, and Hu~Hai.
\newblock Fewclue: A chinese few-shot learning evaluation benchmark.
\newblock \emph{arXiv preprint arXiv:2107.07498}, 2021.

\bibitem[Xue et~al.(2020)Xue, Constant, Roberts, Kale, Al-Rfou, Siddhant,
  Barua, and Raffel]{xue2020mt5}
Linting Xue, Noah Constant, Adam Roberts, Mihir Kale, Rami Al-Rfou, Aditya
  Siddhant, Aditya Barua, and Colin Raffel.
\newblock mt5: A massively multilingual pre-trained text-to-text transformer.
\newblock \emph{arXiv preprint arXiv:2010.11934}, 2020.

\bibitem[Yan et~al.(2022)Yan, Li, Yu, et~al.]{yan2022deep}
Rui Yan, Juntao Li, Zhou Yu, et~al.
\newblock Deep learning for dialogue systems: Chit-chat and beyond.
\newblock \emph{Foundations and Trends{\textregistered} in Information
  Retrieval}, 15\penalty0 (5):\penalty0 417--589, 2022.

\bibitem[Zelikman et~al.(2022)Zelikman, Wu, Mu, and Goodman]{zelikman2022star}
Eric Zelikman, Yuhuai Wu, Jesse Mu, and Noah Goodman.
\newblock Star: Bootstrapping reasoning with reasoning.
\newblock \emph{Advances in Neural Information Processing Systems},
  35:\penalty0 15476--15488, 2022.

\bibitem[Zeng et~al.(2022)Zeng, Liu, Du, Wang, Lai, Ding, Yang, Xu, Zheng, Xia,
  et~al.]{zeng2022glm}
Aohan Zeng, Xiao Liu, Zhengxiao Du, Zihan Wang, Hanyu Lai, Ming Ding, Zhuoyi
  Yang, Yifan Xu, Wendi Zheng, Xiao Xia, et~al.
\newblock Glm-130b: An open bilingual pre-trained model.
\newblock \emph{arXiv preprint arXiv:2210.02414}, 2022.

\bibitem[Zeng et~al.(2021)Zeng, Ren, Su, Wang, Liao, Wang, Jiang, Yang, Wang,
  Zhang, et~al.]{zeng2021pangu}
Wei Zeng, Xiaozhe Ren, Teng Su, Hui Wang, Yi~Liao, Zhiwei Wang, Xin Jiang,
  ZhenZhang Yang, Kaisheng Wang, Xiaoda Zhang, et~al.
\newblock Pangu-$alpha$: Large-scale autoregressive pretrained chinese language
  models with auto-parallel computation.
\newblock \emph{arXiv preprint arXiv:2104.12369}, 2021.

\bibitem[Zhang \& Sennrich(2019)Zhang and Sennrich]{zhang2019root}
Biao Zhang and Rico Sennrich.
\newblock Root mean square layer normalization.
\newblock \emph{Advances in Neural Information Processing Systems}, 32, 2019.

\bibitem[Zhang et~al.(2023{\natexlab{a}})Zhang, Qian, Liu, Heinecke, Meng, Liu,
  Yu, Wang, Savarese, and Xiong]{zhang2023dialogstudio}
Jianguo Zhang, Kun Qian, Zhiwei Liu, Shelby Heinecke, Rui Meng, Ye~Liu, Zhou
  Yu, Huan Wang, Silvio Savarese, and Caiming Xiong.
\newblock Dialogstudio: Towards richest and most diverse unified dataset
  collection for conversational ai, 2023{\natexlab{a}}.

\bibitem[Zhang \& Li(2021)Zhang and Li]{zhang2021commentary}
Min Zhang and Juntao Li.
\newblock A commentary of gpt-3 in mit technology review 2021.
\newblock \emph{Fundamental Research}, 1\penalty0 (6):\penalty0 831--833, 2021.

\bibitem[Zhang et~al.(2023{\natexlab{b}})Zhang, Dong, Li, Zhang, Sun, Wang, Li,
  Hu, Zhang, Wu, et~al.]{zhang2023instruction}
Shengyu Zhang, Linfeng Dong, Xiaoya Li, Sen Zhang, Xiaofei Sun, Shuhe Wang,
  Jiwei Li, Runyi Hu, Tianwei Zhang, Fei Wu, et~al.
\newblock Instruction tuning for large language models: A survey.
\newblock \emph{arXiv preprint arXiv:2308.10792}, 2023{\natexlab{b}}.

\bibitem[Zhang et~al.(2022)Zhang, Roller, Goyal, Artetxe, Chen, Chen, Dewan,
  Diab, Li, Lin, et~al.]{zhang2022opt}
Susan Zhang, Stephen Roller, Naman Goyal, Mikel Artetxe, Moya Chen, Shuohui
  Chen, Christopher Dewan, Mona Diab, Xian Li, Xi~Victoria Lin, et~al.
\newblock Opt: Open pre-trained transformer language models.
\newblock \emph{arXiv preprint arXiv:2205.01068}, 2022.

\bibitem[Zhang et~al.(2021)Zhang, Gu, Han, Chen, Xiao, Sun, Yao, Qi, Guan, Ke,
  et~al.]{zhang2021cpm}
Zhengyan Zhang, Yuxian Gu, Xu~Han, Shengqi Chen, Chaojun Xiao, Zhenbo Sun, Yuan
  Yao, Fanchao Qi, Jian Guan, Pei Ke, et~al.
\newblock Cpm-2: Large-scale cost-effective pre-trained language models.
\newblock \emph{AI Open}, 2:\penalty0 216--224, 2021.

\bibitem[Zhang et~al.(2023{\natexlab{c}})Zhang, Zhang, Li, and
  Smola]{zhang2023automatic}
Zhuosheng Zhang, Aston Zhang, Mu~Li, and Alex Smola.
\newblock Automatic chain of thought prompting in large language models.
\newblock In \emph{The Eleventh International Conference on Learning
  Representations (ICLR 2023)}, 2023{\natexlab{c}}.

\bibitem[Zheng et~al.(2019)Zheng, Huang, and Sun]{zheng-etal-2019-chid}
Chujie Zheng, Minlie Huang, and Aixin Sun.
\newblock {C}h{ID}: A large-scale {C}hinese {ID}iom dataset for cloze test.
\newblock In \emph{Proceedings of the 57th Annual Meeting of the Association
  for Computational Linguistics}, pp.\  778–787. Association for
  Computational Linguistics, 2019.

\bibitem[Zheng et~al.(2023)Zheng, Chiang, Sheng, Zhuang, Wu, Zhuang, Lin, Li,
  Li, Xing, Zhang, Gonzalez, and Stoica]{zheng2023judging}
Lianmin Zheng, Wei-Lin Chiang, Ying Sheng, Siyuan Zhuang, Zhanghao Wu, Yonghao
  Zhuang, Zi~Lin, Zhuohan Li, Dacheng Li, Eric.~P Xing, Hao Zhang, Joseph~E.
  Gonzalez, and Ion Stoica.
\newblock Judging llm-as-a-judge with mt-bench and chatbot arena, 2023.

\bibitem[Zhou et~al.(2023)Zhou, Liu, Xu, Iyer, Sun, Mao, Ma, Efrat, Yu, Yu,
  et~al.]{zhou2023lima}
Chunting Zhou, Pengfei Liu, Puxin Xu, Srini Iyer, Jiao Sun, Yuning Mao, Xuezhe
  Ma, Avia Efrat, Ping Yu, Lili Yu, et~al.
\newblock Lima: Less is more for alignment.
\newblock \emph{arXiv preprint arXiv:2305.11206}, 2023.

\bibitem[Zhou et~al.(2020)Zhou, Zheng, Huang, Huang, and Zhu]{kdconv}
Hao Zhou, Chujie Zheng, Kaili Huang, Minlie Huang, and Xiaoyan Zhu.
\newblock Kdconv: A chinese multi-domain dialogue dataset towards multi-turn
  knowledge-driven conversation.
\newblock In \emph{Proceedings of the 58th Annual Meeting of the Association
  for Computational Linguistics}, pp.\  7098–7108. Association for
  Computational Linguistics, 2020.

\bibitem[Ziang~Leng \& Li(2023)Ziang~Leng and Li]{luotuo}
Qiyuan~Chen Ziang~Leng and Cheng Li.
\newblock Luotuo: An instruction-following chinese language model, lora tuning
  on llama.
\newblock \url{https://github.com/LC1332/Chinese-alpaca-lora}, 2023.

\end{thebibliography}

\clearpage
\appendix

\section{Instruction Template}
\label{sec:instruction_template}


The task instruction prompts for evaluation are provided here:


Test prompt example for MMLU: 
\begin{questionbox}
\textit{\textbf{Context:}} \\
\textcolor{gray}{\textit{(Examplar)}} \\
Question: Which of the following occurred first during the separation of the elements of Pangaea through continental drift? Options: A. Gondwana and Laurasia were formed. B. Africa separated from South America. C. India collided with Eurasia to form the Himalayan mountain chain. D. Australia separated from the rest of the continental landmasses. Answer:A
\\
\textcolor{gray}{\ldots\textit{(Other examplars, if any)}} \\
\\
\textcolor{gray}{\textit{(Test case)}} \\
Question: Experiments on song development in birds have shown that when a young male reared in isolation hears only the song of a different bird species, he will develop an adult song repertoire that lacks certain characteristics typical of his own species. This result shows that the song of his species is most likely Options: A. entirely learned during development B. entirely instinctive C. both instinctive and learned D. dependent upon hormones for proper development Answer: \\
\end{questionbox}
\begin{answerbox}
\textit{\textbf{Response:}} A
\end{answerbox}

\vspace{0.5cm}

Test prompt example for CMMLU: 
\begin{questionbox}
\textit{\textbf{Context:}} \\
\begin{CJK}{UTF8}{gbsn}
\textcolor{gray}{\textit{(Instruction)}} \\
以下是关于(大学教育学)的单项选择题，请直接给出正确答案的选项。\\
\\
\textcolor{gray}{\textit{(Examplar)}} \\
题目：在古代文献记载中，我国西周时期设在王都的小学和大学，总称为() \\
A. 都学 B. 乡学 C. 官学 D. 国学 \\
答案是：D \\
\\
\textcolor{gray}{\ldots\textit{(Other examplars, if any)}} \\
\\
\textcolor{gray}{\textit{(Test case)}} \\
以下是关于(大学教育学)的单项选择题，请直接给出正确答案的选项。\\
题目：教育的本质特征是 () \\
A. 系统性 B. 知识性 C. 科学性 D. 育人性 \\
答案是：
\end{CJK}
\end{questionbox}
\begin{answerbox}
\textit{\textbf{Response:}} D
\end{answerbox}

\clearpage

Test prompt example for C-Eval: 
\begin{questionbox}
\textit{\textbf{Context:}} \\
\begin{CJK}{UTF8}{gbsn}
\textcolor{gray}{\textit{(Instruction)}} \\
以下是关于(中国语言文学)的单项选择题，请直接给出正确答案的选项。\\
\\
\textcolor{gray}{\textit{(Examplar)}} \\
题目：元朝政府曾经实行残酷的民族政策，把全国人民分为\_\_\_\_四个等级。 \\
A. 色目人、蒙古人、汉人、南人 B. 蒙古人、汉人、南人、色目人 C. 蒙古人、南人、色目人、汉人 D. 蒙古人、色目人、汉人、南人 \\
答案是：D \\
\\
\textcolor{gray}{\ldots\textit{(Other examplars, if any)}} \\
\\
\textcolor{gray}{\textit{(Test case)}} \\
以下是关于(中国语言文学)的单项选择题，请直接给出正确答案的选项。\\
题目：《国语》和\_\_\_\_，都是国别史。 \\
A. 《左传》 B. 《战国策》 C. 《史记》 D. 《汉书》\\
答案是：
\end{CJK}
\end{questionbox}
\begin{answerbox}
\textit{\textbf{Response:}} D
\end{answerbox}

\vspace{0.5cm}

Test prompt example for BBH: 
\begin{questionbox}
\textit{\textbf{Context:}} \\
\textcolor{gray}{\textit{(Examplar)}} \\
not ( True ) and ( True ) is \\
Answer: False \\
\\
\textcolor{gray}{\ldots\textit{(Other examplars, if any)}} \\
\\
False or not not not False and True is
\end{questionbox}
\begin{answerbox}
\textit{\textbf{Response:}} True
\end{answerbox}

\vspace{0.5cm}

Test prompt example for $En\Rightarrow Zh$ Machine Translation:
\begin{questionbox}
\textit{\textbf{Context:}} \\
\begin{CJK}{UTF8}{gbsn}
\underline{将以下中文翻译成英文，并输出英文翻译：} \\
\end{CJK}
Local authorities are warning residents in the vicinity of the plant to stay indoors, turn off air-conditioners and not to drink tap water.
\end{questionbox}
\begin{answerbox}
\textit{\textbf{Response:}} \\
\begin{CJK}{UTF8}{gbsn}
当地政府警告核电站附近的居民，要待在室内，关掉空调，不要喝自来水。
\end{CJK}
\end{answerbox}

Test prompt example for $Zh\Rightarrow En$ Machine Translation: 
\begin{questionbox}
\textit{\textbf{Context:}} \\
\begin{CJK}{UTF8}{gbsn}
\underline{将以下英文翻译成中文，并输出中文翻译：} \\
当地政府警告核电站附近的居民，要待在室内，关掉空调，不要喝自来水。
\end{CJK}
\end{questionbox}
\begin{answerbox}
\textit{\textbf{Response:}} \\
Local government warns residents near nuclear power plant to stay indoors, turn off air conditioning, and do not drink bottled water.
\end{answerbox}

\clearpage

Test prompt example for BoolQ: 
\begin{questionbox}
\textit{\textbf{Context:}} \\
Parity (mathematics) -- In mathematics, parity is the property of an integer's inclusion in one of two categories: even or odd. An integer is even if it is evenly divisible by two and odd if it is not even. For example, 6 is even because there is no remainder when dividing it by 2. By contrast, 3, 5, 7, 21 leave a remainder of 1 when divided by 2. Examples of even numbers include -4, 0, 82 and 178. In particular, zero is an even number. Some examples of odd numbers are -5, 3, 29, and 73. \\
\underline{question:} can an odd number be divided by an even number? \\
\underline{answer:}
\end{questionbox}
\begin{answerbox}
\textit{\textbf{Response:}}
yes
\end{answerbox}

\vspace{0.5cm}

Test prompt example for RTE: 
\begin{questionbox}
\textit{\textbf{Context:}} \\
Yet, we now are discovering that antibiotics are losing their effectiveness against illness. Disease-causing bacteria are mutating faster than we can come up with new antibiotics to fight the new variations. \\
\underline{Can we say the following?} \\
Bacteria is winning the war against antibiotics. \\
\underline{OPTIONS: - yes - no}
\end{questionbox}
\begin{answerbox}
\textit{\textbf{Response:}}
yes
\end{answerbox}

\vspace{0.5cm}

Test prompt example for ReCoRD: 
\begin{questionbox}
\textit{\textbf{Context:}} \\
Tracy Morgan hasn't appeared on stage since the devastating New Jersey crash that nearly ended his life last summer, but all that will change this fall when he returns to host Saturday Night Live. NBC announced on Twitter Monday that Morgan, an SNL alum with seven seasons as a cast member under his belt, will headline the third episode of Season 41 airing October 17. For Morgan, 46, it will be a second time hosting the long-running variety show, the first since the June 2014 pileup on the New Jersey Turnpike that killed his friend and mentor James 'Jimmy Mack' McNair. \\
@highlight \\
Morgan, 46, will host third episode of season 41 of SNL airing October 17 \\
@highlight \\
He tweeted to his fans: 'Stoked to be going home...\#SNL' \\
@highlight \\
For the SNL alum who had spent seven years as cast member, it will be a second time hosting the show \\
@highlight \\
Morgan has been sidelined by severe head trauma suffered in deadly June 2014 crash on New Jersey Turnpike that killed his friend \\
@highlight \\
First episode of new SNL season will be hosted by Miley Cyrus, followed by Amy Schumer \\
\end{questionbox}
\begin{answerbox}
\textit{\textbf{Response:}} \\
On October 10, acclaimed comedian and star of the summer box office hit Trainwreck Amy Schumer will make her SNL debut, followed by Morgan a week later.
\end{answerbox}

\clearpage

Test prompt example for WSC: 
\begin{questionbox}
\textit{\textbf{Context:}} \\
Bernard , who had not told the government official that he was less than 21 when he filed for a homestead claim, did not consider that he had done anything dishonest. Still, anyone who knew that he was 19 years old could take his claim away from him . ``him'' \underline{refer to what?}
\end{questionbox}
\begin{answerbox}
\textit{\textbf{Response:}}
anyone
\end{answerbox}

\vspace{0.5cm}

Test prompt example for WiC: 
\begin{questionbox}
\textit{\textbf{Context:}} \\
An emerging professional class. \\
Apologizing for losing your temper, even though you were badly provoked, showed real class. The word `class' used in the same way in the two sentences above, \underline{True or False?} \\
\underline{answer:}
\end{questionbox}
\begin{answerbox}
\textit{\textbf{Response:}}
False
\end{answerbox}

\vspace{0.5cm}

Test prompt example for CoPA: 
\begin{questionbox}
\textit{\textbf{Context:}} \\
The man turned on the faucet, so \_\_ \\
A. The toilet filled with water. B. Water flowed from the spout. \\
answer:
\end{questionbox}
\begin{answerbox}
\textit{\textbf{Response:}}
B
\end{answerbox}

\vspace{0.5cm}

Test prompt example for MultiRC:
\begin{questionbox}
\textit{\textbf{Context:}} \\
\underline{Please read the following article and judge whether the answer to the question is correct:}  What causes a change in motion? The application of a force. Any time an object changes motion, a force has been applied. In what ways can this happen? Force can cause an object at rest to start moving. Forces can cause objects to speed up or slow down. Forces can cause a moving object to stop. Forces can also cause a change in direction. In short, forces cause changes in motion. The moving object may change its speed, its direction, or both. We know that changes in motion require a force. We know that the size of the force determines the change in motion. How much an objects motion changes when a force is applied depends on two things. It depends on the strength of the force. It also depends on the objects mass. Think about some simple tasks you may regularly do. You may pick up a baseball. This requires only a very small force. \\
\underline{questions:} Would the mass of a baseball affect how much force you have to use to pick it up? \\
\underline{answer:} No. \\
\underline{Is this answer True or False?}
\end{questionbox}
\begin{answerbox}
\textit{\textbf{Response:}}
False
\end{answerbox}

\clearpage

Test prompt example for $AX_b$:
\begin{questionbox}
\textit{\textbf{Context:}} \\
\underline{Read the sentence below and answer the question:} The cat sat on the mat. \\
\underline{Question:} The cat did not sit on the mat. \underline{True or False?} \\
\underline{Answer:} \\
\end{questionbox}
\begin{answerbox}
\textit{\textbf{Response:}}
False
\end{answerbox}

\vspace{0.5cm}

Test prompt example for $AX_g$
\begin{questionbox}
\textit{\textbf{Context:}} \\
\underline{Read the sentence below and answer the question:} The taxpayer met with the accountant to get help filing his taxes. \\
\underline{Question:} The accountant sought help filing taxes. \underline{True or False?} \\
\underline{Answer:} \\
\end{questionbox}
\begin{answerbox}
\textit{\textbf{Response:}}
False
\end{answerbox}

\vspace{0.5cm}

Test prompt example for BELEBELE:
\begin{questionbox}
\textit{\textbf{Context:}} \\
\underline{Please read the following article and answer the questions:} \\
Make sure your hand is as relaxed as possible while still hitting all the notes correctly - also try not to make much extraneous motion with your fingers. This way, you will tire yourself out as little as possible. Remember there's no need to hit the keys with a lot of force for extra volume like on the piano. On the accordion, to get extra volume, you use the bellows with more pressure or speed. \\ 
\underline{Please answer the following multiple-choice questions:} \\
According to the passage, what would not be considered an accurate tip for successfully playing the accordion? Options: (A) For additional volume, increase the force with which you hit the keys (B) Keep unnecessary movement to a minimum in order to preserve your stamina (C) Be mindful of hitting the notes while maintaining a relaxed hand (D) Increase the speed with which you operate the bellows to achieve extra volume  \underline{Answer: (} \\
\end{questionbox}
\begin{answerbox}
\textit{\textbf{Response:}}
B
\end{answerbox}

\vspace{0.5cm}
 
Test prompt example for CLTS:
\begin{questionbox}
\textit{\textbf{Context:}} \\
\begin{CJK}{UTF8}{gbsn}
\underline{简要概括以下文本：} \\
温布尔登网球锦标赛官方微博发布消息称，全英草地网球俱乐部今天非常遗憾地决定，由于新冠状病毒疫情持续蔓延产生的公共卫生安全问题，将取消2020年温布尔登网球锦标赛，第134届锦标赛将于2021年6月28日至7月11日举行。（原题为《2020年温网锦标赛取消将于2021年6月至7月举行》）(本文来自澎湃新闻，更多原创资讯请下载“澎湃新闻”APP) \\
\underline{概括为：}
\end{CJK}
\end{questionbox}
\begin{answerbox}
\textit{\textbf{Response:}} \\
\begin{CJK}{UTF8}{gbsn}
由于新冠状病毒疫情持续蔓延产生的公共卫生安全问题，将取消2020年温布尔登网球锦标赛，第134届锦标赛将于2021年6月28日至7月11日举行。
\end{CJK}
\end{answerbox}

\vspace{1.0cm}

Test prompt example for OpenBA-Chat:
\begin{questionbox}
\textit{\textbf{Context:}} \\
\begin{CJK}{UTF8}{gbsn}
Human: 你叫什么名字啊? \\
Assistant: 我叫OpenBA,是一款由苏州大学的自然语言处理实验室开发的大语言模型。 \\
Human: 你是GPT架构的模型吗? \\
Assistant: 我不基于GPT架构开发,而是基于浅编码器-深解码器的架构开发的模型。 \\
Human: 你会哪些语言? \\
Assistant: \\
\end{CJK}
\end{questionbox}
\begin{answerbox}
\textit{\textbf{Response:}} \\
\begin{CJK}{UTF8}{gbsn}
在预训练及微调数据中,我学习了中英两种语言。正如我的名字OpenBA,我是一个双语(Bilingual)的模型。
\end{CJK}
\end{answerbox}

\vspace{0.5cm}

Test prompt example for OpenBA-InstructGen:
\begin{questionbox}
\textit{\textbf{Context:}} \\
\begin{CJK}{UTF8}{gbsn}
Please generate the instruction according to the context I provide: \\
Nice to meet you! I am an AI assistant, how can I help you today? \\
\end{CJK}
\end{questionbox}
\begin{answerbox}
\textit{\textbf{Response:}} \\
\begin{CJK}{UTF8}{gbsn}
Hello, nice to meet you!
\end{CJK}
\end{answerbox}

\vspace{0.5cm}

Test prompt example for OpenBA-Tool:
\begin{questionbox}
\textit{\textbf{Context:}} \\
\begin{CJK}{UTF8}{gbsn}
Who wrote i got a brand new pair of rollerskates? \\
\end{CJK}
\end{questionbox}
\begin{answerbox}
\textit{\textbf{Response:}} \\
\begin{CJK}{UTF8}{gbsn}
WikiSearch("Brand New Key") $\rightarrow$ People will take it any way they want to take it." In a 2013 interview with music journalist Ray Shasho, Melanie revealed the true origin of "Brand New Key": Of course I can see it symbolically with the key, but I just thought of roller skating. I was fasting with a twenty seven day fast on water. I broke the fast and went back to my life living in New Jersey and we were going to a flea market around six in the morning. On the way back ...and I had just broken the fast, from the flea market, we]Melanie
\end{CJK}
\end{answerbox}

\section{Chinese Flan Collection}
\label{appdix:chinese_flan_collection}

\begin{figure}[H]
    \centering
    \begin{questionbox}
    \textit{\textbf{Example of Unified Json Format:}} \\
    \{  \\
        \begin{CJK}{UTF8}{gbsn}
        \quad "instruction": "请将输入的英语句子翻译成中文", \\
        \quad "input": "Yet not nearly enough has been invested in this effort.", \\
        \quad "output": "但目前这方面的投入还远远不够。", \\
       \quad "domain": "多领域", \\
        \quad "task": "机器翻译"  \\
        \end{CJK}
    \} \\
    \end{questionbox}
    \caption{An example of the unified format of Chinese Flan Dataset.}
    \label{fig:unified-format}
\end{figure}

As shown in Fig.~\ref{fig:unified-format}, the unified format of each data includes "instruction," "input," "output," "domain" and "task", where "instruction" denotes the description of the task that provides LLM with a clear purpose. 
"input" and "output" are the question and answer respectively. 
"domain" is the topic of the task, such as medicine, news, etc. 
"task" indicates the type categorized into one of eighteen task types. 
Table~\ref{tab:chinese_flan} shows all the tasks and the source of the instruction datasets in each task.

\begin{table*}[ht]
\centering
\renewcommand{\arraystretch}{1.5}
\resizebox{0.95\textwidth}{!}{
    \begin{tabular}{c p{13cm}<\centering c}
        \toprule
        \bf Task & \bf Source & \bf Dataset \\
        \midrule
        
        \multirow{4}{*}{Question Answering} & 
        ~\cite{XQA} & XQA \\
        & ~\cite{duan2016nlpcc} & ChineseDBQA \\
        & ~\cite{Artetxe:etal:2019} & Xquad \\
        & \url{https://www.luge.ai/\#/luge/dataDetail?id=40} & ChineseBiomedicalQA \\
        \midrule
        
        \multirow{5}{*}{Text Classification} & 
        ~\cite{conneau2018xnli} & XNLI \\
        & \url{http://tcci.ccf.org.cn/conference/2014/dldoc/evtestdata6.zip} & Chinese News Categorization \\
        & \url{https://tianchi.aliyun.com/dataset/133838?spm=a2c22.28136470.0.0.6e5a6a23SPZMrX&from=search-list} & TNEWS \\
        & \url{https://huggingface.co/datasets/dirtycomputer/ChnSentiCorp_htl_all} & ChnSentiCorp\_htl\_all \\
        & \url{https://storage.googleapis.com/cluebenchmark/tasks/iflytek_public.zip} & iflytek \\
        \midrule
        
        \multirow{4}{*}{Sentiment Classification} & 
        ~\cite{FewCLUE} & FewCLUE EPRSTMT \\
        & \url{https://www.luge.ai/\#/luge/dataDetail?id=25} & ChnSentiCorp \\
        & \url{https://www.heywhale.com/mw/dataset/5e09a9eb2823a10036b126c0/file} & BDCI 2019 \\
        & \url{https://www.luge.ai/\#/luge/dataDetail?id=20} & NLPCC14-SC \\
        \midrule
        
        \multirow{2}{*}{Named Entity Recognition} & \url{https://huggingface.co/datasets/msra_ner} & MSRA\_NER \\
        & \url{https://storage.googleapis.com/cluebenchmark/tasks/cluener_public.zip} & CLUE Fine-Grain NER \\
        \midrule
        
        \multirow{5}{*}{Text Matching} & 
        ~\cite{xu-etal-2020-clue} & CLUE WSC 2020 \\
        & ~\cite{xu-etal-2020-clue} & CMNLI \\
        & ~\cite{OCNLI} & OCNLI \\
        & \url{https://www.luge.ai/\#/luge/dataDetail?id=39} & CINLID \\
        & \url{https://tianchi.aliyun.com/dataset/106411} & AFQMC \\
        \midrule
        
        \multirow{3}{*}{Text Summarization} & ~\cite{CSL} & CSL \\
        & ~\cite{ladhak-wiki-2020} & WikiLingua Chinese \\
        & \url{http://tcci.ccf.org.cn/conferen} & Weibo Oriented Chinese News Summarization NLPCC2015 \\
        \midrule
        
        \multirow{3}{*}{Reading Comprehension} & ~\cite{xu-etal-2020-clue} & C3 \\
        & ~\cite{cui-emnlp2019-cmrc2018} & CMRC2018 \\
        & ~\cite{DRCD} & DRCD \\
        \midrule
        
        \multirow{2}{*}{Question Generation} & 
        ~\cite{DuReader} & DuReader\_QG \\
        & \url{https://tianchi.aliyun.com/dataset/dataDetail?dataId=86895} & TCM Literature Question Generation \\
        \midrule
        
        \multirow{3}{*}{Dialogue} & 
         ~\cite{douban} & douban \\
        & ~\cite{kdconv} & kdconv \\
        & \url{https://www.luge.ai/\#/luge/dataDetail?id=38} & Chinese Persona Chat \\
        \midrule
        
        \multirow{4}{*}{Machine Translation} & \cite{duh18multitarget} & mttt \\
        & ~\cite{bright_xu_2019_3402023} & translation 2019 zh \\
        & ~\cite{WMT19} & WMT19 en-zh \\
        & \url{https://www.kaggle.com/datasets/garyongguanjie/wikititles-zhen} & wikititles\_en-zh \\
        \midrule
        
        Cloze Test & ~\cite{zheng-etal-2019-chid} & ChiD \\
        \midrule
        
        Text Generation & ~\cite{AdvertiseGen} & AdvertiseGen \\
        \midrule
        
        Semantic Analysis & ~\cite{Math23K} & Math23K \\
        \midrule
        
        Relation Extraction & ~\cite{DuIE2.0} & DuIE2.0 \\
        \midrule
        
        Grammatical Error Correction & ~\cite{MD-SCS} & MD-SCS \\
        \midrule
        
        Fact-checking & ~\cite{CHEF} & CHEF \\
        \midrule
        
        Interpretable Evaluation & ~\cite{DuExplain} & DuExplain \\
        \midrule
        
        Event Extraction & \url{https://tianchi.aliyun.com/dataset/dataDetail?dataId=110904} & tianchi\_event\_doclevel\_attr \\
        
        \midrule
    \end{tabular}
}

\caption{All types of tasks and the source of the instruction datasets in each task.}
\label{tab:chinese_flan}
\end{table*}


\end{document}